\documentclass{article}
\usepackage{mdframed}
\usepackage{xcolor}
\usepackage{hyperref}
\usepackage{enumitem}
\usepackage[table]{xcolor}
\newcommand{\std}[1]{{\scriptsize\color{gray}$\pm#1$}}
 
\hypersetup{
    colorlinks=true,
    linkcolor=StageOneBlue,  % Links for Figures/Sections
    urlcolor=StageOneBlue,   % URLs
    citecolor=StageOneBlue,  % Citations (the [1], [2] links)
}
\usepackage{url}
\definecolor{HypothesisBG}{RGB}{254, 240, 217}       % Light orange background
\definecolor{HypothesisBorder}{RGB}{215, 135, 40} 

\definecolor{StageThreePink}{RGB}{255,105,180} % Example pink color, adjust as needed
\definecolor{StageTwoOrange}{RGB}{255,165,0} % Example orange color, adjust as needed
\definecolor{StageOneBlue}{RGB}{70,130,180} % Example blue color, adjust as needed
\definecolor{TakeawayPinkBG}{HTML}{FAE7F1}      % A light, soft pink for the background
\definecolor{TakeawayPinkBorder}{HTML}{C0397B} 

\definecolor{modelblue}{RGB}{0, 110, 180}
\definecolor{physred}{RGB}{180, 50, 50}
\definecolor{errorgray}{RGB}{100, 100, 100}
\usepackage[most]{tcolorbox}

\newtcolorbox[auto counter]{takeawaypink}[1][]{
    enhanced,
    attach boxed title to top left={yshift=-2mm, xshift=3mm},
    colback=white,
    colframe=HypothesisBorder,
    colbacktitle=HypothesisBorder,
    coltitle=black,
    fonttitle=\bfseries,
    boxrule=1pt,
    arc=4pt,
    breakable,
    title={Takeaway~\thetcbcounter}, 
    #1
}
\definecolor{denblue}{RGB}{179,205,227}   % soft blue
\definecolor{natorange}{RGB}{253,180,98}  % soft orange

\definecolor{curvgreen}{HTML}{C6F0C2}

\usepackage{tikz}
\usetikzlibrary{shapes.geometric, arrows.meta, positioning, decorations.pathmorphing, shapes.misc,shadows, calc, fadings}

\definecolor{PhysicsGrey}{RGB}{45, 52, 54}
\definecolor{SSLTeal}{RGB}{0, 184, 148}
\definecolor{CollapseRed}{RGB}{214, 48, 49}
\tikzfading[name=fade out, inner color=transparent!0, outer color=transparent!100]

\usepackage[most]{tcolorbox}
\usepackage{fontawesome5}
\tcbuselibrary{skins}
\tcbuselibrary{breakable}

\definecolor{FinetuneGreenBG}{HTML}{EAF6EA}   % Light green background
\definecolor{FinetuneGreenBorder}{HTML}{2CA02C} % Matplotlib 'C2'

\newtcolorbox{hypothesis}[1][]{
    enhanced,
    attach boxed title to top left={yshift=-2mm, xshift=3mm},
    colback=white,
    colframe=HypothesisBorder,
    colbacktitle=HypothesisBorder,
    coltitle=black,
    fonttitle=\bfseries,
    boxrule=1pt,
    arc=4pt,
    breakable,
    title={\faLightbulb \quad Hypothesis}, % The title is now static
    #1
}

\definecolor{TakeawayGreenBG}{HTML}{E0F2E9} % A soft, light green for the background
\definecolor{TakeawayGreenBorder}{HTML}{006D2C} % A complementary dark green for the border and text
\usepackage[most]{tcolorbox}

\newtcolorbox{boxtheorem}{
    enhanced,
    colback=StageOneBlue!10,      % 10% Blue background (Very soft)
    colframe=StageOneBlue,        % Solid StageOneBlue border
    boxrule=0.5pt,                % Thin, elegant border
    arc=2pt,                      % Rounded corners
    left=5pt, right=5pt, top=5pt, bottom=5pt, % Inner padding
    breakable                     % Allows splitting across pages
}

\usepackage{microtype}
\usepackage{graphicx}
\usepackage{siunitx}
\usepackage{subcaption}
\usepackage{booktabs} % for professional tables
\usepackage{amsthm}

\usepackage[preprint]{icml2026}

\usepackage{amsmath}
\usepackage{amssymb}
\usepackage{mathtools}
\usepackage{amsthm}

\usepackage[capitalize,noabbrev]{cleveref}

\theoremstyle{plain}

\theoremstyle{definition}

\theoremstyle{remark}

\usepackage[textsize=tiny]{todonotes}

\icmltitlerunning{The Observer Effect in World Models}

\usepackage[most]{tcolorbox}

\begin{document}

\twocolumn[
 \icmltitle{The Observer Effect in World Models:\\ Invasive Adaptation Corrupts Latent Physics}

 % \icmltitle{Self-Supervised Learning Can Encode Physics Linearly;\\ Downstream Adaptation Can Corrupt It}
 % \icmltitle{Validating Physical World Models via Non-Invasive Linear Probing}
 % Non-Invasive Probing Reveals Intrinsic Physical World Models
 % \icmltitle{Validating World Models Without Modifying Them\\ via Non-Invasive Probing}
  %\icmltitle{Non-Invasive Probing Reveals Linear Physical Invariants,\\ Invasive Finetuning Corrupt Them}



% 1. Define symbols
% (*) for equal contribution
\icmlsetsymbol{equal}{*}
\icmlsetsymbol{advise}{$\dagger$}

\begin{icmlauthorlist}
    % First two authors: Equal Contribution
    \icmlauthor{Christian Internò}{yyy,dbf,equal}
    \icmlauthor{Jumpei Yamaguchi}{comp,dbf,equal}
    
    \icmlauthor{Loren Amdahl-Culleton}{sch}
    \icmlauthor{Markus Olhofer}{dbf,advise}
    % Last two authors: Co-advising
    \icmlauthor{David Klindt}{scqkh,advise}
    \icmlauthor{Barbara Hammer}{yyy,advise}
\end{icmlauthorlist}

% 2. Define affiliations (ensure tags match identifiers above exactly)
\icmlaffiliation{yyy}{Bielefeld University}
\icmlaffiliation{comp}{Tokyo Institute of Technology}
\icmlaffiliation{sch}{Simplex, Astera Institute}
\icmlaffiliation{dbf}{Honda Research Institute EU}
\icmlaffiliation{scqkh}{Cold Spring Harbor Laboratory}

% 3. Define corresponding author
\icmlcorrespondingauthor{Christian Internò}{christian.interno@uni-bielefeld.de}

  \vskip 0.2in
]

% 2. Print the notice
% We manually combine the standard equal contribution macro with our custom co-advising text
\printAffiliationsAndNotice{\icmlEqualContribution. \textsuperscript{$\dagger$}Co-advising.}

\begin{abstract}
Determining whether neural models internalize physical laws as world models, rather than exploiting statistical shortcuts, remains challenging, especially under out-of-distribution (OOD) shifts. Standard evaluations often test latent capability via downstream adaptation (e.g., fine-tuning or high-capacity probes), but such interventions can change the representations being measured and thus confound what was learned during self-supervised learning (SSL). We propose a non-invasive evaluation protocol, \textit{PhyIP}. We test whether physical quantities are linearly decodable from frozen representations, motivated by the \textit{linear representation hypothesis} \citep{nanda2023emergentlinearrepresentationsworld}. Across fluid dynamics and orbital mechanics, we find that when SSL achieves low error, latent structure becomes linearly accessible. \textit{PhyIP} recovers internal energy and Newtonian inverse-square scaling on OOD tests (e.g., $\rho > 0.90$). In contrast, adaptation-based evaluations can collapse this structure ($\rho \approx 0.05$). These findings suggest that adaptation-based evaluation can obscure latent structures and that low-capacity probes offer a more accurate evaluation of physical world models.
\end{abstract}

\section{Introduction}
\label{sec:intro}
AI for natural sciences has evolved from computational acceleration to the construction of ``World Models'' \cite{1207631, LeCun2022APT}, aiming to synthesize vast observational data into representations that encode the governing physical laws of the system~\cite{wang2023scientific, bommasani2022opportunitiesrisksfoundationmodels}, rather than just statistical shortcuts~\citep{geirhos2020shortcut}. 
\begin{hypothesis}
 \label{hyp:Hypothesy}

\vspace{0.2em}
\centering
\resizebox{1.0\linewidth}{!}{%
\begin{tikzpicture}[
    font=\sffamily\bfseries,
    >=Stealth,
    scale=1.0,
    line cap=round,
    line join=round,
    % --- Custom Styles ---
    source/.style={
        draw=black!80,
        line width=1.5pt,
        top color=black!20,
        bottom color=white,
        drop shadow={opacity=0.4, shadow xshift=2pt, shadow yshift=-2pt}
    },
    ssl_beam/.style={
        top color=StageOneBlue!60,
        bottom color=StageOneBlue!10,
        opacity=0.4
    },
    ssl_manifold/.style={
        draw=StageOneBlue!100,
        line width=2.0pt,
        top color=StageOneBlue!30,
        bottom color=StageOneBlue!5,
        drop shadow={opacity=0.4, color=StageOneBlue!50!black}
    },
    ft_beam/.style={
        top color=StageTwoOrange!60,
        bottom color=StageTwoOrange!10,
        opacity=0.4
    },
    ft_manifold/.style={
        draw=StageTwoOrange!100,
        line width=2.0pt,
        top color=StageTwoOrange!30,
        bottom color=StageTwoOrange!5,
        decorate,
        decoration={random steps, segment length=3mm, amplitude=1.5mm},
        drop shadow={opacity=0.4, color=StageTwoOrange!50!black}
    }
] % <-- THIS ] MUST EXIST

% Optional: make randomness deterministic so it doesn't change each compile
% \pgfmathsetseed{123}

    % 1. PHYSICS (Top Center)
    \begin{scope}[shift={(0, 2.8)}]
        \node[text=black!90, scale=1.5] at (0, 1.6) {Physical System };

        % The Source Blob
        \draw[source, smooth cycle, tension=0.7]
          plot coordinates {(-3.5, 0.2) (-1.5, 1.0) (2.0, 0.8) (3.5, -0.2) (0, -0.8)};

        % --- VECTOR FIELD VISUAL ---
        \begin{scope}[shift={(0, 0)}, opacity=0.6]
            \clip[smooth cycle, tension=0.7]
              plot coordinates {(-3.5, 0.2) (-1.5, 1.0) (2.0, 0.8) (3.5, -0.2) (0, -0.8)};

            \foreach \x in {-3.0, -1.8, -0.6, 0.6, 1.8, 3.0}
              \foreach \y in {-0.2, 0.4}
                \draw[->, black!70, thick]
                  ({\x + rand*0.2}, {\y + rand*0.1}) -- ++(0.4, 0.15);
        \end{scope}

        \coordinate (Source) at (0, -0.8);
    \end{scope}

    % 2. SSL SIDE (Left)
    \begin{scope}[shift={(-4.0, 0)}]
        \node[text=StageOneBlue!80!black, scale=1.5] at (-0.5, 1.8) {SSL ($m_\theta$)};

        % Beam
        \fill[ssl_beam] (Source) -- (-2.8, -0.3) -- (2.8, -0.3) -- cycle;

        \foreach \i in {1,...,60}
          \fill[StageOneBlue!90!black] ({rand*1.8}, {rand*1.0 + 1.2}) circle (1.5pt);

        % --- SHADOW (Drawn BEHIND) ---
        \begin{scope}[shift={(0, -2.0)}]
            \fill[black!10] (0,0) ellipse (2.5 and 0.6);
            \begin{scope}
                \clip (0,0) ellipse (2.5 and 0.6);
                \draw[black!40, thick] (-3,-1) grid[step=0.4] (3,1);
            \end{scope}
            \node[text=black!70, scale=1.1, anchor=north] at (0, -0.7) {Dynamics Recovered Linearly};

            \draw[dotted, thick, black!30] (-2.0, 1.4) -- (-2.0, 0);
            \draw[dotted, thick, black!30] (2.2, 1.2) -- (2.2, 0);
        \end{scope}

        % --- MANIFOLD (Drawn ON TOP) ---
        \begin{scope}[shift={(0, -0.4)}]
            \draw[ssl_manifold, smooth cycle, tension=0.6]
              plot coordinates {(-2.5, 0.8) (-1.0, 1.5) (2.5, 1.0) (2.2, -1.0) (-2.0, -0.8)};
            \begin{scope}
                \clip[smooth cycle, tension=0.6]
                  plot coordinates {(-2.5, 0.8) (-1.0, 1.5) (2.5, 1.0) (2.2, -1.0) (-2.0, -0.8)};
                \draw[StageOneBlue!80, very thick] (-3,-2) grid[step=0.5, xslant=0.2] (3,3);
            \end{scope}
        \end{scope}
    \end{scope}

    % 3. FT SIDE (Right)
    \begin{scope}[shift={(4.0, 0)}]
        \node[text=StageTwoOrange!80!black, scale=1.5] at (0.5, 1.8) {$m_{\theta'}$};

        % Beam
        \fill[ft_beam] (Source) -- (-1.2, -0.3) -- (1.2, -0.3) -- cycle;
        \node[star, star points=5, fill=StageTwoOrange!90!black, scale=0.9] at (0, 1.2) {};
        \node[star, star points=5, fill=StageTwoOrange!90!black, scale=0.9] at (-0.4, 0.8) {};

        % --- SHADOW (Drawn BEHIND) ---
        \begin{scope}[shift={(0, -2.0)}]
            \fill[black!10] (0,0) ellipse (2.5 and 0.6);

            \foreach \i in {1,...,40}
              \fill[black!30] ({rand*2.0}, {rand*0.5}) circle (1.8pt);

            \node[text=black!70, scale=1.1, anchor=north] at (0, -0.7) {Dynamics Lost};

            \draw[dotted, thick, black!30] (-2.0, 1.4) -- (-2.0, 0);
            \draw[dotted, thick, black!30] (2.0, 1.4) -- (2.0, 0);
        \end{scope}

        % --- MANIFOLD (Drawn ON TOP) ---
        \begin{scope}[shift={(0, -0.4)}]
            \draw[ft_manifold] (0,0) ellipse (2.2 and 1.2);

            \foreach \i in {1,...,12}
              \draw[StageTwoOrange!90!black, very thick]
                ({rand*1.5}, {rand*0.6}) -- ++({rand*0.4}, {rand*0.4});
        \end{scope}
    \end{scope}

    % --- ARROW SECTION ---
    \draw[->, >=Stealth, line width=2.5pt, black!50, dashed]
        (-1.1, -0.8) -- (1.5, -0.8)
        node[midway, above=8pt, scale=1.4, font=\bfseries, text=black!60] {$\theta \to \theta'$}
        node[midway, below=8pt, scale=0.9, font=\sffamily\bfseries, align=center, text=black!60]
        {Downstream task\\adaptation};

\end{tikzpicture}%
} % <-- closes resizebox

\end{hypothesis}

The aspiration is that a neural network, trained on diverse physical regimes, will implicitly learn the corresponding physical laws, analogous to the historical transition from describing motion (kinematics) to uncovering the forces that drive it (dynamics)~\citep{science1165893}.

However, despite high predictive fidelity within training distributions, neural models often fail to capture universal physical mechanisms~\citep{coveney2025aineedsphysicsphysics, motamed2025generativevideomodelsunderstand}. If a model cannot distinguish universal laws from correlations, its scientific utility remains limited~\citep{chen2022automated, wang2023scientific}. 

The prevailing methodology for adapting these models to new downstream tasks is 'pretrain-then-fine-tune', where the backbone is updated alongside a randomly initialized head for a specific downstream task~\citep{wortsman2021robust, NEURIPS2024_f573c364}.  Recently, this paradigm has been extended to validate intrinsic knowledge. Invasive adaptation and high-capacity nonlinear probes are used to test whether inductive biases in models align with a postulated world model~\citep{vafa2025foundationmodelfoundusing, belinkov2022probing}. However, theoretical works on probing warn that such high-capacity interventions often ``learn the task'' themselves rather than extracting it~\citep{hewitt2019designinginterpretingprobescontrol, ravichander2021probing}, while feature distortion dynamics work suggests that, during adaptation, noisy gradients warp the backbone to fit random initializations rather than the task structure~\citep{kumar2022finetuning, trivedi2023a}.

This degrades physical representations, challenging the \textit{Linear Representation Hypothesis} (LRH)~\citep{nandaemergent, DICARLO2007333},  which posits that features are encoded as linear directions in the activation space of capable models.  Furthermore, when modeling continuous physical evolution (ranging from orbital ordinary differential equations (ODEs) to fluid partial differential equations (PDEs)), the structural equivalence between residual networks and Euler discretizations becomes critical~\citep{NEURIPS2018_69386f6b, Haber_2018}. Since the model effectively learns to act as a numerical integrator for these dynamics~\citep{chen2018neural}, invasive fine-tuning can disrupt the weights into non-physical regimes.

Bridging the gap between predictive fidelity and the evaluation of physical understanding requires a shift towards interpretable probing frameworks~\citep{bereska2024mechanisticinterpretabilityaisafety} and rigorous experimental design and control~\cite{hewitt2019designinginterpretingprobescontrol, belinkovprobing}. We argue that apparent failures to encode physics are often not failures of learning but artifacts of the measurement process itself (i.e., the downstream adaptation task), which can distort the underlying representation~\cite{santi2025flow, belinkov2022probing}.

We propose the \textbf{Non-Invasive Physical Probe} (\textit{PhyIP}), which is a mechanistic evaluation framework that uses time-independent linear readouts on frozen SSL representations to extract latent conserved quantities. These quantities are then distilled into interpretable formulas via symbolic regression (SR)~\citep{biggio2021neural} and validated against strict control baselines for probes~\citep{hewitt-liang-2019-designing}. These controls are designed to rule out possible false positives and false negatives under Out-Of-Distribution (OOD) settings.

Consequently, we posit that adopting neural dynamics models as a \textit{fixed measurement instrument}~\citep{Jaeger2001TheechoST, mencattini2026exploratorycausalinferencesaence} is vital for valid scientific inquiry. In classical experimental design, the measuring tool should remain invariant relative to the subject to avoid confounding the observation with the instrument's own adaptation~\citep{Mari2023}. 

We validate this hypothesis on high-fidelity simulations from the TheWell benchmark~\citep{howard2023well}: a \textit{2D Turbulent Radiative Layer}~\citep{stachenfeld2021learned}, a \textit{3D Red Supergiant}~\citep{Goldberg_2022}, and a \textit{3D Supernova Explosion}~\citep{hirashima2023surrogate}. We find that low-error SSL models actively encode physical dynamics into linear subspaces across diverse regimes. In radiative turbulence, \textit{PhyIP} recovers the internal energy law ($E \approx 1.5P$) with high precision. In the more complex \textit{Red Supergiant} setting, it spontaneously develops a correction term for convective kinetic energy ($\sim \rho v_r^2$). 

Conversely, we show that invasive methods such as non-linear probes~\cite{belinkov2022probing}, last-layer fine-tuning (LLFT)~\cite{kirichenko2023last}, and full fine-tuning via Inductive Bias Probes (IBP)~\cite{vafa2025foundationmodelfoundusing} can systematically mislead. In the supernova simulation, these methods report high accuracy despite SSL failure. Furthermore, in replicating the orbital mechanics experiment of \citet{vafa2025foundationmodelfoundusing}, we provide a mechanistic analysis of this destructive intervention. We observe a collapse in representational similarity (CKA)~\citep{kornblith2019similarity} in deep blocks, where the optimizer minimizes loss by discarding time-varying features (speed, radius) and relying instead on constants (mass). This shows that physical knowledge was latent in the SSL model but corrupted by the measurement process itself, analogous to the \textit{``observer effect''}~\cite{SassolideBianchi2013, heisenberg1927}.

\noindent \textbf{Our contributions:}
\begin{enumerate}[leftmargin=*, noitemsep, nosep]
    \item \textbf{Non-Invasive Physical Probe (PhyIP):} We introduce a framework to extract latent physical quantities from frozen representations without inducing distortion.
    \item \textbf{Experimental Design:} We derive a bound linking SSL error ($\epsilon$) and functional curvature ($K_{\Phi}$) to linear decodability, enabling strict experimental control.
    \item \textbf{Physics Recovery:} We successfully recover fundamental laws, including internal energy ($E \approx 1.5P$) and gravitational force ($F \propto 1/r^2$), across complex fluid and orbital benchmarks where invasive methods fail.
    \item \textbf{Invasive Corruption Evidence:} These results demonstrate that adaptation acts as a destructive intervention that overwrite internal world models. 
\end{enumerate}

\section{Preliminaries \& Framework}
\label{sec:pre}

To distinguish between a model that encodes physics and one that memorizes data, we must formalize the interaction between the physical dynamics, the neural architecture, and the probe as ``measurement instrument''.

\textbf{Data Generating Process (DGP):}
We assume the physical system is a time-dependent field $u(z, t)$ on a domain $\Omega \subseteq \mathbb{R}^D$ evolving via PDEs: $\frac{\partial u}{\partial t} = \mathcal{F}(u, \nabla u, \dots)$. To align with the discrete nature of computation, this continuous field is discretized into a finite-dimensional state vector $x(t) \in \mathcal{X} \subseteq \mathbb{R}^n$ (we also write \(x_t\) with a subscript to denote the functional dependence on time). Consequently, the evolution follows an autonomous ODE: $ \dot{x}(t) = f(x(t))$ where $f: \mathcal{X} \to \mathbb{R}^n$ is the Lipschitz continuous vector field approximating the continuous dynamics.

\textbf{Observational Data:}
The continuous DGP is observed at discrete time intervals $\Delta t$, yielding a dataset of trajectories $\mathcal{T}$. A trajectory $\tau \in \mathcal{T}$ is a sequence of states $(x_0, x_1, \dots, x_T)$ where $x_{t+1} = x_t + \int_{t}^{t+\Delta t} f(x(s))\, ds$.
The learning task is to approximate this transition operator.

\textbf{Physical Observables \& Curvature:}
Let $\Phi: \mathcal{X} \to \mathbb{R}^k$ be a target physical functional defining the quantity of interest $s = \Phi(x)$. We assume $\Phi$ is $C^2$-smooth and define its \textit{Curvature} $K_\Phi$ as the Lipschitz constant of the gradient $\nabla \Phi$ with respect to the Euclidean norm:
\begin{equation}
    \| \nabla \Phi(x_a) - \nabla \Phi(x_b) \|_2 \le K_\Phi \|x_a - x_b\|_2
\end{equation}
where $x_a, x_b \in \mathcal{X}$. $K_\Phi$ quantifies non-linearity in the input space. By Taylor's theorem, the deviation of $\Phi$ from its linear tangent is at most $\frac{1}{2}K_\Phi \| \Delta x\|^2$.  For linear functionals (e.g., Linear Momentum), $K_\Phi = 0$. For non-linear interactions (e.g., Gravitational Forces), $K_\Phi > 0$.

\textbf{Neural Dynamics Model:}
Let $m_\theta$ be a model that maps an input trajectory $x_{0:t}$ to a latent representation $h_t$ and a next-step prediction $\hat{x}_{t+1}$. The model is trained to minimize the self-supervised prediction error :
\begin{equation}
\epsilon = \mathbb{E} [ \mathcal{L}_{\text{SSL}}(\hat{x}_{t+1}, x_{t+1}) ]   
\end{equation}
Where $\mathcal{L}_{\text{SSL}}$ is typically the Mean Squared Error (MSE), quantifying the model's ability to approximate the underlying physical transition operator.

\textbf{Probe:}
A probe is a diagnostic function $P_W: \mathbb{R}^d \to \mathbb{R}^k$ with learnable parameters $W$. It maps the \textit{frozen} latent representation $h_t$ to the value of the target physical quantity:
\begin{equation}
s_{t+1} = \Phi(x_t).
\end{equation}
\subsection{Why a Linear Probe for Physics?}
\label{sec:theory}
While the \textit{Linear Representation Hypothesis}~\citep{DICARLO2007333} is well-documented for neural models trained on language data~\citep{park2023the, nandaemergent}, this property remains under-explored in models trained on physical systems. \textit{We hypothesize that the linear encoding of physical dynamics is an emergent capability of a successful self-supervised optimization task.}

Most effective scientific models (e.g., Transformers, U-Nets, and Fourier Neural Operators) effectively model continuous dynamics via an Euler-like discretization~\citep{Haber_2018, chen2018neural}. 
We formalize this \textit{incremental residual prediction} property as follows: the model predicts a future state as an additive update: the model predicts a future state as an additive update $\hat{x}_{t+1} = x_t + g(h_t)$, where $g$ is a learned decoder representing the state displacement.
Complementarily, an \textit{Origin-Preserving Decoder}: $g(\mathbf{0}) = \mathbf{0}$ ensures that a null latent state ($\mathbf{h}_t = \mathbf{0}$) corresponds to the identity map (no physical update), a property actively encouraged by zero-initialization~\citep{goyal2017accurate} and weight decay~\citep{he2015deepresiduallearningimage}.

A model $m_\theta$ with low error $\epsilon$ must maintain a representation $h$ that is locally consistent with the physical update $\Delta x$. We define the optimal linear probe $P_{W^*}$ as the best first-order approximation mapping this latent space to the target quantity update.
The expected probe error in recovering the evolution of the physical target quantity $\Delta s = s(t+1) - s(t)$ is bounded by: 
% \begingroup
% \setlength{\abovedisplayskip}{7pt}
% \setlength{\belowdisplayskip}{2pt}
\begin{equation}
\label{eq:bound}
\begin{split}
\underbrace{\mathbb{E} \left[ \left| P_{W^*} h_t - \Delta s \right|^2 \right]}_{\text{Probe Error}} \le & \underbrace{C_1 \cdot \epsilon}_{\text{Modeling SSL Error}} \\
+ \underbrace{C_2 \left[ K_{\Phi}^2 \cdot\text{Var}(x) \right]}_{\text{Curvature Error}} + \underbrace{\mathcal{O}(\Delta t^4)}_\text{Discretization Error}
\end{split}
\end{equation}
% \begin{equation}
% \label{eq:bound}
% \begin{split}
% \underbrace{\mathbb{E} \left[ \left| P_{W^*} h_t - \Delta s \right|^2 \right]}_{\text{Probe Error}} \le & \underbrace{C_1 \cdot \epsilon}_{\text{Modeling SSL Error}} \\
% + \underbrace{C_2 \left[ K_{\Phi}^2 \cdot\text{Var}(x) \right]}_{\text{Curvature Error}} + \underbrace{\mathcal{O}(\Delta t^4)}_\text{Discretization Error}
% \end{split}
% \end{equation}
% \endgroup
See \Cref{app:proof} for derivation.
This inequality establishes the linear probe as a rigorous diagnostic tool. The error decomposes into two sources: \textbf{1)} Modeling SSL Error ($\epsilon$): If the model fails to predict the dynamics (high $\epsilon$), the probe fails. \textbf{2)} Curvature ($K_\Phi$): If the physical dynamics is highly non-linear, a linear approximation naturally suffers.

Crucially, physical dynamics can be non-linear; for instance, in orbital mechanics, the force vector $\nabla \Phi(x)$, where \(x\) is a position trajectory, rotates as the planet moves. While a time-dependent probe (one that ``rotates'' its weights to match local gradients) might achieve lower error, it risks approximating the physics via its own capacity rather than measuring the representation~\citep{hewitt2019designinginterpretingprobescontrol}. Therefore, the probe optimal $W^*$ must be a single constant matrix across all time steps. A probe success under these constraints serves as a possible \textit{litmus test} for evaluation.

\begin{takeawaypink}[title=Takeaway 1: Probes as Fixed Instruments]
\textit{The measurement instrument must be a \textbf{fixed, time-invariant linear operator} targeting the next state. Consequently, a successful readout under this strict condition serves as evidence that the SSL model has transformed the complex non-linear dynamics into a linearizable representation, consistent with the error bound derived in Eq.~\ref{eq:bound}.}
\end{takeawaypink}

\subsection{\textit{PhyIP}: Non-Invasive Probe for Physics}
\label{sec:method}
\begin{figure*}[h]
\centering
\includegraphics[width=\linewidth]{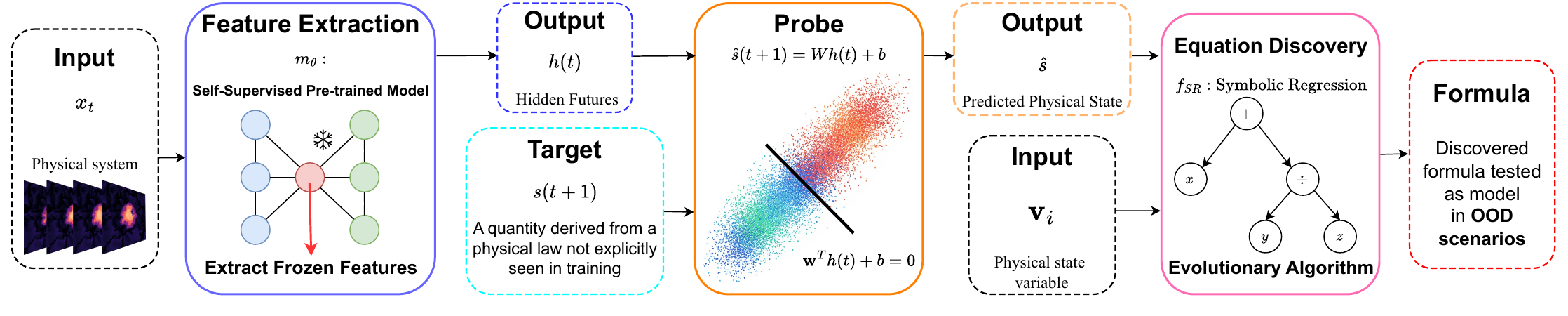}
\caption{\textbf{Overview of \textit{PhyIP} framework.} 
\textcolor{StageOneBlue}{\textbf{(1) Feature Extraction:}} We extract frozen activations, $h(t)$ from a SSL model. 
\textcolor{StageTwoOrange}{\textbf{(2) Linear Probing:}} A linear probe is trained to predict a new physical quantity, $\hat{s}(t+1) = Wh(t) + b$. Its success on OOD tests indicates that $\hat{s}$ is linearly encoded. 
\textcolor{StageThreePink}{\textbf{(3) Equation Discovery \& Validation:}} SR translates the probe's function into a symbolic formula, $\hat{\Phi}_{SR}$, and tests it on OOD data to confirm its physical plausibility.}
    \label{fig:methodology}
\end{figure*}
The objective of our framework is to probe the internal geometry of a pre-trained SSL model $m_{\theta}$ on continuous physical trajectories without corrupting its learned representation. This three-stage process is summarized in Figure~\ref{fig:methodology}.

\textbf{Feature Extraction and Linear Probing.}
To probe the geometry of the activation space, $\mathcal{H} \subseteq \mathbb{R}^d$, we train a linear probe, $P_W: \mathcal{H} \to \mathcal{Q}$, where $\mathcal{Q} \subseteq \mathbb{R}^k$ is the space of the physical quantity. The target dimension $k \ge 1$ depends on the quantity being probed; for example, $k=1$ for a scalar (such as force magnitude) or $k=2$ for a 2D vector (such as the force $\vec{F} = (F_x, F_y)$). 
Here, $h(t)$ is the internal activation vector (e.g., from the decoder block) of the pre-trained SSL model, $m_{\theta}$. This activation is the result of the model processing \Cref{sec:theory}, $h(t)$ represents the model's internal ``plan'' to execute the update. Following \Cref{sec:theory}, we probe $\Delta s_t$. The target is the physical state at the next time step, $\hat{s}(t+1)$, via the transformation:
\begin{equation}
    \hat{s}(t+1) = Wh(t; m_\theta) + b
    \end{equation}
$\hat{s}(t+1) \in \mathcal{Q} \subseteq \mathbb{R}^k$ is the predicted quantity. The linear mapping matrix $W \in \mathbb{R}^{k \times d}$ and the bias vector $b \in \mathbb{R}^k$ transform the high-dimensional representation $h(t) \in \mathcal{H} \subseteq \mathbb{R}^d$ into the low-dimensional space $\mathcal{Q}$.

The probe's optimal parameters, $(W^*, b^*)$, are found by minimizing a loss function (e.g., MSE) on a training set $D_{\text{train}} = \{(h_i, s)\}_{i=1}^N$. Here, the index $i$ iterates over the $N$ samples in the dataset and does not represent the time $t$. Each sample $i$ is a pair created from a specific time step $t$ in the simulation, such that $h_i = h(t)$ (the model activation at time $t$) and $s_{i} = s(t+1)$ (the ground-truth physical quantity at the \textit{next} time step):
\begingroup
\setlength{\abovedisplayskip}{4pt}
\setlength{\belowdisplayskip}{4pt}
\begin{equation}
(W^*, b^*) = \arg\min_{W, b} \frac{1}{N} \sum_{i=1}^{N} \|s_{i} - (W h_i + b)\|^2
\end{equation}
\endgroup
This loss function minimizes the Euclidean distance between the ground-truth quantity $s_{i}$ and the probe's linear prediction $\hat{s}_{i}$. The optimization process is constrained to learn only $W$ and $b$ (the linear map parameters), while the complex, nonlinear feature extraction of the underlying model $m_{\theta}$ (which produces $h_i$) remains frozen. 

The probe's success is then quantified by its generalization performance on an (OOD) test set ($D_{\text{OOD}}$). For our physical systems, $D_{\text{OOD}}$ consists of simulations where key generative parameters (e.g., central star mass, gravitational constant, initial velocity, or boundary conditions) are sampled outside the distribution used for the self-supervised pre-training. A success here confirms that the linear encoding is a robust physical invariant, not a memorized correlation.

\textbf{Equation Discovery.}
While an OOD-generalizing probe confirms a meaningful geometry, its learned mapping $W$ remains opaque. To decode this geometry, we employ Symbolic Regression (SR). This step treats the probe's output as the ground truth. The SR algorithm is given the physical state variables $x_i$ (e.g., position, pressure) as inputs and the probe's predictions $\hat{s}_{i}$ as targets.
SR searches a space of symbolic expressions $\mathcal{G}$ for an optimal formula $\hat{\Phi}_{SR}^*$:
\begingroup
\setlength{\abovedisplayskip}{4pt}
\setlength{\belowdisplayskip}{4pt}
\begin{equation}
    \hat{\Phi}_{SR}^* = \arg\min_{\hat{\Phi} \in \mathcal{G}} \frac{1}{N} \sum_{i=1}^{N} \| \hat{s}_{i} - \hat{\Phi}(x_i) \|^2
\end{equation}

\endgroup
To ensure $\hat{\Phi}_{SR}^*$ represents a true physical principle, we treat it as a standalone physical hypothesis and evaluate its zero-shot generalization on $D_{\text{OOD}}$. A low loss on unseen simulation parameters serves as robust validation that the original model $m_{\theta}$ successfully encoded the governing law.

\section{Probing Complex Fluid Dynamics Systems}
\label{sec:fluid}
\begin{table*}[t]
\centering
\small
\setlength{\tabcolsep}{2.5pt} 
\caption{\textbf{Comprehensive Probe Analysis on Fluid Dynamics.} OOD evaluation of internal energy recovery. PhyIP (top) reliably extracts physical laws from frozen representations when the SSL error ($\epsilon$) is low, recovering the ideal gas law ($E \approx 1.5P$) and kinetic corrections ($\rho v_r^2$). The SN-3D results demonstrate that invasive probes (bottom) can hide collapse, whereas PhyIP correctly identifies it.}
\label{tab:merged_fluid_results}

% Resizebox ensures the table fits within textwidth
\resizebox{\textwidth}{!}{
\begin{tabular}{l | ccc | ccc | ccc}
\toprule
% 1. VISUALIZATIONS
& \multicolumn{3}{c|}{\includegraphics[width=5cm, height=3cm]{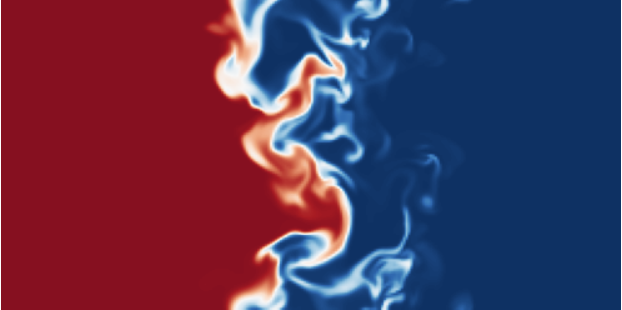}} 
& \multicolumn{3}{c|}{\includegraphics[width=5cm, height=3cm]{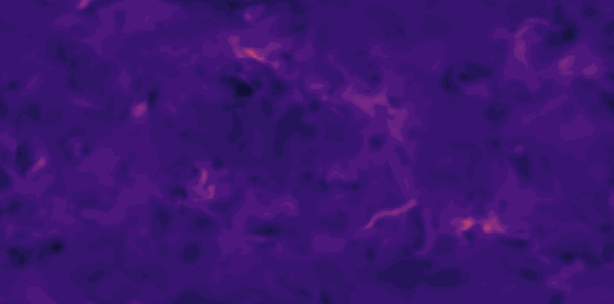}} 
& \multicolumn{3}{c}{\includegraphics[width=4cm, height=3cm,]{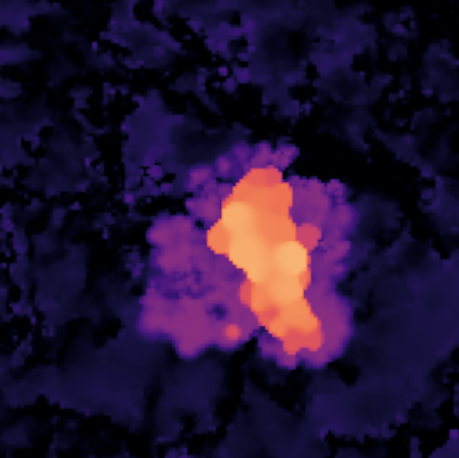}} \\

% 2. DATASET NAMES
& \multicolumn{3}{c|}{\textbf{2D Turbulent Layer} ($N=9$ OOD)} 
& \multicolumn{3}{c|}{\textbf{3D Red Supergiant} ($N=3$ OOD)} 
& \multicolumn{3}{c}{\textbf{3D Supernova} ($N=27$ OOD)} \\
\midrule

% 3. TASK GROUPING
& \multicolumn{2}{c|}{\bfseries Probe Task} & {\bfseries SSL Task}
& \multicolumn{2}{c|}{\bfseries Probe Task} & {\bfseries SSL Task}
& \multicolumn{2}{c|}{\bfseries Probe Task} & {\bfseries SSL Task} \\
\cmidrule{2-3} \cmidrule{4-4} \cmidrule{5-6} \cmidrule{7-7} \cmidrule{8-9} \cmidrule{10-10}

% 3. METRICS HEADER
\textbf{Model / Method} 
& MAPE $\downarrow$ & $\rho \uparrow$ & $\boldsymbol{\epsilon_{OOD}} \downarrow$ 
& MAPE $\downarrow$ & $\rho \uparrow$ & $\boldsymbol{\epsilon_{OOD}} \downarrow$ 
& MAPE $\downarrow$ & $\rho \uparrow$ & $\boldsymbol{\epsilon_{OOD}} \downarrow$ \\
\midrule

% --- PART 1: ORIGINAL ARCHITECTURE COMPARISON ---
\multicolumn{10}{l}{\textit{\textbf{I. \textit{PhyIP}~(\Cref{sec:method})}}} \\
\textbf{UNetConvNext} 
& \bfseries 36.9 \std{1.2} & \bfseries 0.83 \std{0.02} & \bfseries 0.20 \std{0.01}
& \bfseries 18.2 \std{0.9} & \bfseries 0.91 \std{0.01} & \bfseries 0.02 \std{0.00}
& 140.4 \std{12.1} & 0.15 \std{0.05} & \bfseries 0.30 \std{0.02} \\

\textbf{UNetClassic}  
& 42.3 \std{2.1} & 0.71 \std{0.04} & 0.26 \std{0.02}
& 25.6 \std{1.5} & 0.69 \std{0.03} & 0.09 \std{0.01}
& 135.9 \std{10.5} & 0.08 \std{0.02} & 0.40 \std{0.03} \\

\textbf{FNO}            
& 76.0 \std{5.3} & 0.61 \std{0.06} & 0.49 \std{0.04}
& 95.1 \std{4.2} & 0.22 \std{0.08} & 0.05 \std{0.01}
& 95.3 \std{8.1}  & 0.18 \std{0.04} & 0.36 \std{0.02} \\

\textbf{TFNO}           
& 89.7 \std{6.1} & 0.67 \std{0.05} & 0.58 \std{0.05}
& 92.5 \std{5.0} & 0.25 \std{0.07} & 0.04 \std{0.01}
& \bfseries 92.1 \std{7.8}  & \bfseries 0.21 \std{0.03} & 0.36 \std{0.03} \\

\midrule
% --- PART 2: BASELINE ANALYSIS ---
\multicolumn{10}{l}{\textit{\textbf{II. Baselines \& Probes}}} \\

(i) Linear Probe on \textbf{Raw Inputs}
& 58.5 \std{4.1} & 0.32 \std{0.05} & - 
& 88.2 \std{3.5} & 0.25 \std{0.04} & - 
& 142.1 \std{15.0} & 0.22 \std{0.01} & - \\

(ii) \textbf{Time-Dependent} Probe
& 22.1 \std{1.0} & 0.88 \std{0.02} & -
& 15.4 \std{0.8} & 0.95 \std{0.01} & -
& 93.2 \std{11.5} & 0.64 \std{0.05} & - \\

(iii) \textbf{MLP Probe~\cite{belinkov2022probing}}
& \textcolor{CollapseRed}{37.2} \std{1.5} &\textcolor{CollapseRed}{ 0.72} \std{0.03} & -
& \textcolor{CollapseRed}{19.1} \std{1.2} & \textcolor{CollapseRed}{0.82} \std{0.02} & -
& \textcolor{CollapseRed}{125.5} \std{14.2} & \textcolor{CollapseRed}{0.48} \std{0.06} & - \\

(iv) \textbf{LL-FT~\cite{kirichenko2023layerretrainingsufficientrobustness}}
& \textcolor{CollapseRed}{32.5} \std{2.8} & \textcolor{CollapseRed}{0.80} \std{0.04} & -
& \textcolor{CollapseRed}{23.4} \std{4.1} & \textcolor{CollapseRed}{0.80} \std{0.05} & -
& \textcolor{CollapseRed}{65.0} \std{13.1} & \textcolor{CollapseRed}{0.59} \std{0.04} & - \\

(v) \textbf{Full FT (IBP~\cite{vafa2025foundationmodelfoundusing})}
& \textcolor{CollapseRed}{41.2} \std{3.5} & \textcolor{CollapseRed}{0.81} \std{0.04} & -
& \textcolor{CollapseRed}{21.1} \std{6.4} & \textcolor{CollapseRed}{0.80} \std{0.03} & -
& \textcolor{CollapseRed}{18.3} \std{18.5} & \textcolor{CollapseRed}{0.71} \std{0.01} & - \\

\midrule \midrule

% --- PART 3: ORIGINAL METADATA ---
\textbf{SSL Input Vars} 
& \multicolumn{3}{c|}{$\{\rho, P, v_x, v_y\}$} 
& \multicolumn{3}{c|}{$\{\rho, P, v_r, v_\theta, v_\phi\}$} 
& \multicolumn{3}{c}{$\{\rho, P, T, v_x, v_y, v_z\}$} \\
\addlinespace[4pt]

% PROBE TARGET
\textbf{Probe Target} 
& \multicolumn{3}{c|}{$E_{\text{int}} = \int 1.5 P \, dV \quad (\gamma=5/3)$} 
& \multicolumn{3}{c|}{$E_{\text{int}} = \int \rho u \, dV$} 
& \multicolumn{3}{c}{$E_{\text{int}} = \int \rho u \, dV$} \\
\addlinespace[4pt]

% SYMBOLIC LAW
\textbf{Discovered Law} ($\hat{\Phi}_{SR}$)
& \multicolumn{3}{c|}{ $\mathbf{E \approx 1.48 \cdot P}$} 
& \multicolumn{3}{c|}{ $\mathbf{E \approx 1.45 P + 0.42 \rho v_r^2}$} 
& \multicolumn{3}{c}{ $E \approx 0.35 - \left[ \frac{0.06}{(P + 0.2)} \right]$} \\

\bottomrule
\end{tabular}
} % End resizebox
\end{table*}
\textbf{Setup:} To demonstrate the generality of \textit{PhyIP}, we applied it to three high-dimensional fluid dynamics simulations from the \texttt{TheWell benchmark}~\citep{howard2023well}: a \textit{2D Turbulent Radiative Layer}~\citep{stachenfeld2021learned}, a \textit{3D Red Supergiant Convective Envelope}~\citep{Goldberg_2022}, and a \textit{3D Supernova Explosion}~\citep{hirashima2023surrogate}. The experimental setting satisfies the conditions of \Cref{sec:theory}. 
We tested neural models—including U-Net~\citep{ronneberger2015u}, UNetConvNext~\citep{unet2020}, FNO~\citep{li2021fourier}, and TFNO~\citep{li2021fourierneuraloperatorparametric}—trained solely on self-supervised next-state prediction. 

These models function as residual predictors, satisfying the setting discussed in \cref{sec:pre}.
Using our non-invasive method \textit{PhyIP}, a linear probe was trained on frozen activations (e.g., U-Net \texttt{neck}). To predict the global total internal energy $E_{\text{int}}$, we implemented the probe as a \texttt{$1\times 1$ Convolutional Layer} (kernel size 1) followed by a global summation. This forces the probe to predict the local energy density contribution $(\rho u)_{i,j}$ at each voxel $(i,j)$ using only the local latent vector $h_{i,j}$. The final prediction is the sum over the domain: $\hat{E}_{\text{int}} = \sum_{i,j} \text{Probe}(h_{i,j}) \Delta V$. To distill these probes into interpretable formulas, we restricted the \texttt{PySR} search space to basic arithmetic operators $\{+, -, \times, /\}$ with strict dimensional consistency constraints (penalty $10^{12}$) to prioritize simplicity over curve fitting.

\textbf{Control Baselines \& Invasive Probes.} All probes are trained using supervised pairs derived exclusively from the in-Distribution (ID) SSL training set, ensuring strictly zero-shot evaluation on OOD regimes. We compare against: \textbf{i)} Raw Input Baseline: A linear regression on flattened raw input fields $x_t \in \mathbb{R}^{C \times H \times W}$. \textbf{ii)} Time-Dependent Probe: A linear probe $\{W_t\}$ optimized per time-step removing the curvature penalty ($K_\Phi$) from our bound to quantify ``curvature mismatch.'' \textbf{iii)} Non-Linear MLP Probe~\cite{belinkov2022probing}: A 3-layer MLP $(254\rightarrow32)$ on frozen activations $h_t$. Finally, we test a  \textbf{iv)} Last-Layer Fine-Finetuning (LL-FT)~\cite{kirichenko2023layerretrainingsufficientrobustness} and \textbf{v)} Full Fine-tuning Adaptation (all $\theta$) via the IBP~\cite{vafa2025foundationmodelfoundusing}. Success here despite high SSL error ($\epsilon$) confirms the probe is \textit{learning from scratch} rather than measuring the world model.

\begin{figure*}[ht]
    \centering
    % --- Top Panel: Scatter Plots ---
    \begin{subfigure}{\linewidth}
        \centering
        \includegraphics[width=\linewidth]{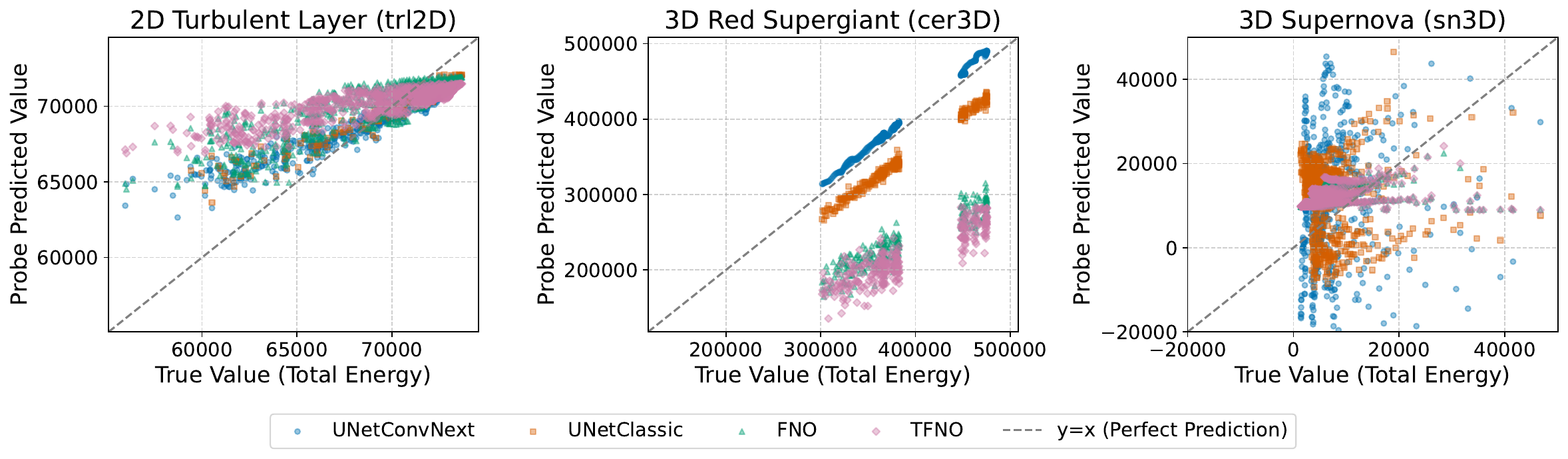}
        \label{fig:combined_scatter}
    \end{subfigure}
    
    \vspace{-1em} % Space between panels

    % --- Bottom Panel: Mean Curves ---
    \begin{subfigure}{\linewidth}
        \centering
        \includegraphics[width=\linewidth]{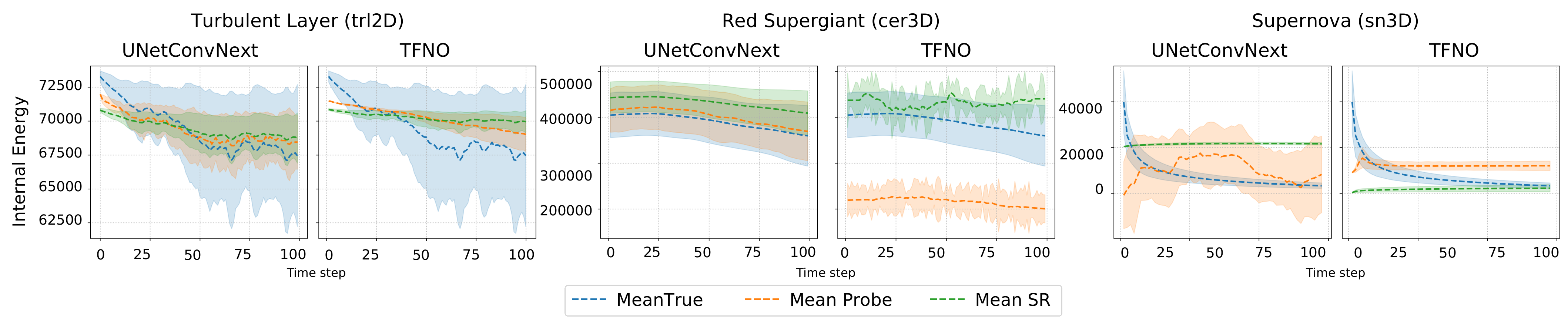}
        \label{fig:combined_curves}
    \end{subfigure}
    \caption{
    \textbf{Probing Latent Physical Laws. (Top)} The Non-Invasive Probe successfully extracts Total Internal Energy from frozen SSL representations in TRL-2D and RSG-3D (linear alignment), but correctly identifies representational collapse in the SN-3D experiment. \textbf{(Bottom)} Zero-shot generalization. Discovered symbolic formulas (Mean SR) accurately predict energy dynamics on unseen simulation parameters for the successful models}
    \label{fig:combined_probe_analysis}
\end{figure*}

% \begin{figure*}[ht]
%     \centering
%     \includegraphics[width=0.9\linewidth]{Figures/comparison_ALL_DATASETS_probe_scatter_grid_all_timesteps.pdf}
%     \caption{\textbf{Probe Prediction vs. True Energy (All Models \& Datasets).} 
%     \textbf{(Left/Center)} For TRL-2D and RSG-3D, the U-Net's predictions form a tight, linear cluster ($\rho \approx 0.9$) corresponding to its low SSL test error. 
%     \textbf{(Right)} For the complex Supernova, all models fail SSL prediction (high $\epsilon$), resulting in scattered, uncorrelated probe outputs.}
%     \label{fig:fluid_scatter_comparison}
% \end{figure*}

% \begin{figure*}[h]
%     \centering
%         % *** This is the plot from fig:trl2d_mean ***
%         \includegraphics[width=\textwidth]{Figures/Our Probe (5).pdf}
%            \caption{\textbf{Mean Energy Prediction vs. Time.} 
%     Each panel plots the mean predicted energy (y-axis) over time (x-axis) for UNetConvNext and TFNO models. (\textit{blue}) si 'Mean True' energy, \textit{green/red}) is the mean probe and mean SR.}
%     \label{fig:mean_curve_comparison}
% \end{figure*}

\textbf{OOD Evaluation.} We validated on 39 Out-of-Distribution (OOD) test sets, 9 unseen cooling rates for TRL-2D, 3 distinct stellar evolution phases for RSG-3D, and 27 novel environments varying ambient gas density and metallicity  for Supernova.
Results summarized in \Cref{tab:merged_fluid_results} reveal a divergence that provides empirical validation for \Cref{sec:theory}.

For the \textbf{2D Turbulent Radiative Layer}, the U-Net architectures minimized the SSL test error ($\epsilon \approx 0.19$) significantly better than FNO models ($\epsilon \approx 0.50$) with the \textit{PhyIP} linear encoding $\rho=0.83$ and MAPE $=36.9$. \textit{PhyIP} symbolic regression recovered the equation $\mathbf{E \approx 1.48 \cdot P}$, which matches the constant ($1.5$ for $\gamma=5/3$) with $<2\%$ error. This success is visually confirmed in \Cref{fig:combined_probe_analysis} (top), where predictions form a linear cluster, and in \Cref{fig:combined_probe_analysis} (bottom), where the symbolic model (green) of UNetConvNext accurately forecasts the energy decay on unseen cooling rates. In contrast, higher-capacity or invasive baselines do not improve upon the linear readout: the MLP probe reaches $\rho=0.72$ (MAPE $=37.2\%$), and both last-layer fine-tuning and full fine-tuning perform slightly worse than \textit{PhyIP} on OOD.

The \textbf{3D Red Supergiant} simulation provides the strongest validation. The UNetConvNext achieved the lowest OOD prediction error ($\epsilon=0.0201$), outperforming the FNO ($\epsilon=0.0505$). As expected, this SSL mastery enabled precise \textit{PhyIP}'s linear decoding ($\rho=0.91$). Here, \textit{PhyIP} shows that rather than just retrieving the static pressure law ($1.5 P$), the probe recovered a composite symbolic expression $\mathbf{E \approx 1.45 P + 0.42 \rho v_r^2}$. We validate this term via dimensional analysis, the quantity $\rho v^2$ has the units of energy density ($J/m^3$), matching the units of Pressure ($P$). This confirms that the term is a valid correction for kinetic energy. As shown in \Cref{fig:combined_probe_analysis} (bottom), this ``convective proxy'' allows the symbolic formula of UNetConvNext to track the ground-truth energy evolution, while the scatter plot (\Cref{fig:combined_scatter}, Center) confirms the precision of the \textit{PhyIP} decoding. \textit{PhyIP} probe reduces MAPE error by nearly $5\times$ compared to the Raw Input Baseline ($\rho = 0.25$ and MAPE $=88.2\%$), proving high-level abstraction. The MLP Probe ($\rho = 0.82$ and MAPE $=19.1\%$) together with LL-FT and IBP fails to improve upon \textit{PhyIP}, verifying that the relevant geometry is linearly encoded. Finally, by using the Time-Dependent Probe we gain better performance ($\rho = 0.95$ and MAPE $=15.4\%$) as expected.

% Update the RSG-3D discussion in Section 5:

Finally, the \textbf{3D Supernova} test proved challenging for all architectures, with SSL prediction error at a high $\epsilon \approx 0.30-0.37$. 
Empirically, no model when probed with \textit{PhyIP} achieved a correlation above $\rho=0.2$. Furthermore, \textit{PhyIP} symbolic regression of the best performer model TFNO failed to find any physical law, fitting instead a spurious rational function $\mathbf{E \approx 0.35 - [\frac{0.06}{(P + 0.2)}]}$ that implies unphysical inverse scaling. This failure is visualized in \Cref{fig:combined_probe_analysis}, where both the probe and symbolic formula completely fail to track the energy decay. \textbf{This failure exposes the danger of invasive probing:} while \textit{PhyIP} probe correctly diagnoses this (MAPE $=140.4\%$), the full fine-tuning via IBP (Table 1, row v) achieves a deceptively low MAPE = $18.3\%$ and $\rho=0.71$. This massive discrepancy ($140\% \to 18\%$ for MAPE and $0.21 \to 0.71$) confirms that the invasive probes did not measure the model's knowledge but rather overwrote it, hallucinating competence where there was none.

% \textbf{Baseline Validation.} A comprehensive baseline analysis (\Cref{tab:merged_fluid_results}) confirms the structural validity of these findings across all regimes. For successful models (TRL, RSG), the linear probe drastically outperforms the raw input baseline (e.g., RSG: 18.2\% vs 88.2\% MAPE), proving active compression, while the failure of the non-linear MLP probe to improve performance (19.1\%) empirically verifies that the learned geometry is indeed linear. Conversely, the Supernova failure case exposes the danger of invasive methods: while our non-invasive probe correctly diagnoses the model's ignorance (140.4\% error), the invasive finetuning probe achieves a deceptively low error (18.3\%). This massive discrepancy confirms that invasive methods do not measure intrinsic knowledge but rather overwrite it, hallucinating competence where none exists.

\begin{takeawaypink}[title=Takeaway 2: Strict Experimental Control]
\textit{
In complex fluids, faithful evaluation requires \textbf{strict experimental control}. \textit{PhyIP} succeeds only when SSL models the dynamics (low OOD $\epsilon$), while invasive adaptation can \emph{mask} backbone failure (SN-3D: $\text{MAPE } 140\% \to 18\%$) by learning the probe task itself and distorting otherwise valid linear structure (as seen in TRL-2D and RSG-3D).
}
\end{takeawaypink}

\section{The Confounding Nature of Invasive Probes}
\label{sec:finetuning}

\begin{figure*}[t] % Change to figure* if you need it to span the full page width
    \centering
    % --- Top Image: Manifold ---
    \includegraphics[width=\linewidth, trim={0 0 0 0}, clip]{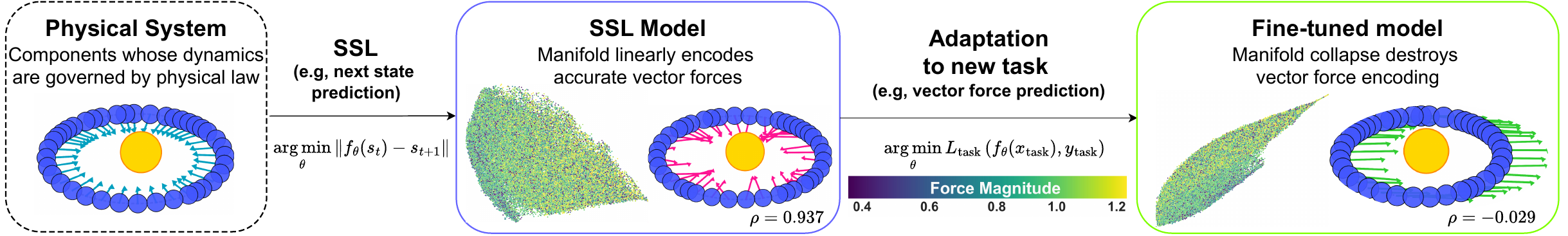}
    
    \vspace{0.1cm} % Pulls the second image up closer. Adjust value as needed.
    
    % --- Bottom Image: Performance ---
    \includegraphics[width=0.9\linewidth, trim={0 0 0 0}, clip]{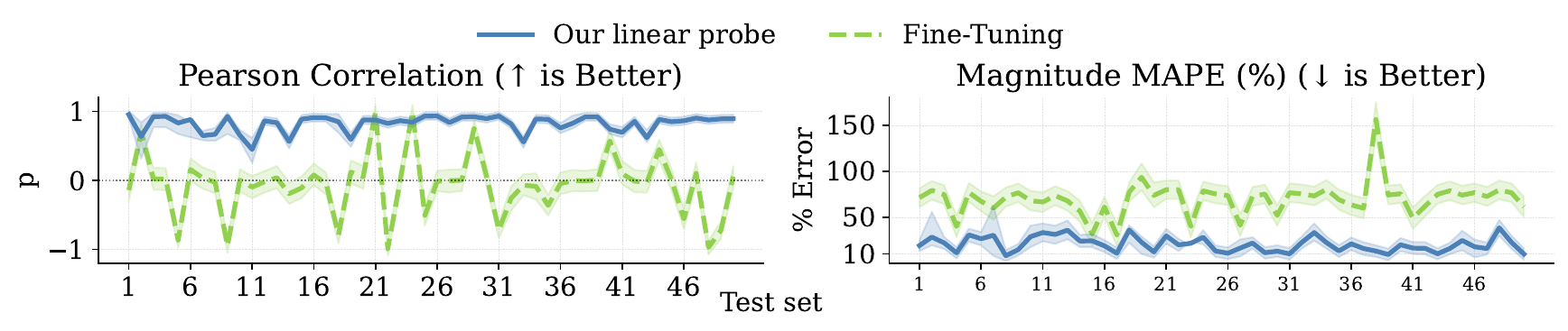}
    
    \vspace{0cm} % Pulls the caption up
    
    \caption{\textbf{The Failure of Invasive Probing.} \textbf{Top:} The orderly geometry of the SSL model (center) is destroyed by invasive fine-tuning (right), dropping correlation from $\rho=0.94 \to -0.03$. \textbf{Bottom:} Quantitative Impact. This geometric destruction causes the invasive probe (Green) to fail erratically on OOD task, whereas our non-invasive probe (Blue) remains robust ($\rho > 0.8$).}
    \label{fig:combined_failure}
\end{figure*}

\textbf{Motivation.} While the Fluid Dynamics experiment (\Cref{sec:fluid}) demonstrate scalability, it is difficult to isolate the exact mechanism of invasive corruption. To provide a controllable test, we replicate the Orbital Mechanics experiment of \citet{vafa2025foundationmodelfoundusing} for the \textit{inductive bias probe}. This setting allows us to: \textbf{(1)} directly compare \textit{PhyIP} against invasive methods in a highly nonlinear task (force vector) and \textbf{(2)} perform a mechanistic analysis to point exactly when and how invasive probes overwrite knowledge.

\textbf{Setup:} A 109M parameter Transformer ($m_{\theta}$), pre-trained (SSL) on orbital trajectories, is subjected to full-parameter fine-tuning on a small datasets where the output is the force vector at each point in the trajectory. The weights of the entire architecture $\theta$ are updated to $\theta'_i$ by minimizing the task loss over a fine-tuning dataset $D_{\text{task},i}$. For each pair $(x, s)$, $x$ represents the input trajectory and $s$ the ground-truth physical target variable, the force vector \begin{equation}
\theta'_i = \arg\min_{\theta} \sum_{(x,s)\in D_{\text{task},i}} \mathcal{L}_{\text{task}}(m_{\theta}(x), s)
\end{equation}
The subsequent inductive bias analysis is performed on the fine-tuned model $m_{\theta'_i}$.

\textbf{Representation Collapse:} We investigate whether invasive fine-tuning acts as a \textit{destructive intervention} that degrades the SSL model's ($m_{\theta}$) latent geometry. To establish a robust baseline, we employ the non-invasive probe \textit{PhyIP}(\Cref{sec:method}) on the frozen activations of the decoder \texttt{blocks.9}.  We compare \textit{PhyIP} against the fully adapted fine-tuned model via IBP. As shown in \Cref{fig:combined_failure}, our probe maintains high performance ($\rho > 0.85$, MAPE $< 30\%$) across 50 OOD tests, while the fine-tuned model exhibits erratic failure with MAPE spiking between $50\%$ and $150\%$ (see Appendix \ref{app:solar_system_analysis} for solar system validation results). 

\Cref{tab:orbital_baselines} confirms the non-triviality of the task: the Raw Input Probe fails (MAPE $65.2\%$). The Time-Dependent probe attains low error (MAPE $12.1\%$ and $\rho = 0.96$) by adapting to local gradients, effectively bypassing the curvature term $K_\Phi$ in \Cref{eq:bound}. Introducing limited adaptation with LL-FT partially mitigates the degradation observed with the IBP (MAPE $45.3\%$, $\rho = 0.65$), but it still performs substantially worse than \textit{PhyIP} (MAPE $24.5\%$, $\rho = 0.91$).

\begin{table}[t]
\centering
\small
\setlength{\tabcolsep}{3pt} % Tighten padding for single column fit
\caption{\textbf{Orbital Mechanics Baseline Analysis.} Comparison of our Non-Invasive Probe against baselines on the 50 OOD test set (force vector task).}
\label{tab:orbital_baselines}
\resizebox{\columnwidth}{!}{% Resizes table to exactly fill the column width
\begin{tabular}{l | cc}
\toprule
% HEADER
& \multicolumn{2}{c}{\textbf{OOD test sets}} \\
\cmidrule{2-3}
\textbf{Method} & \textbf{MAPE} $\downarrow$ & $\boldsymbol{\rho}$ \textbf{(Pearson)} $\uparrow$ \\
\midrule

\rowcolor{StageOneBlue!10} 
\textbf{I. Non-Invasive Probe (\textit{PhyIP})} & \textbf{24.5 \std{4.2}} & \textbf{0.91 \std{0.02}} \\

\midrule
% II. BASELINES
\multicolumn{3}{l}{\textit{\textbf{II. Baselines \& Controls}}} \\

(i) Linear Probe on \textbf{Raw Inputs} & 65.2 \std{4.5} & 0.45 \std{0.03} \\

(ii) \textbf{Time-Dependent} Probe & 12.1 \std{1.2} & 0.96 \std{0.01} \\

(iii) \textbf{MLP Probe} & 22.5 \std{3.0} & 0.88 \std{0.04} \\

(iv) \textbf{LL-FT} & 45.3 \std{5.1} & 0.65 \std{0.06} \\

% V. FINETUNING (Inferred from Green Dashed Line)
(v) \textbf{Full FT (IBP)} & \textcolor{CollapseRed}{81.5 \std{12.4}} & \textcolor{CollapseRed}{0.05 \std{0.35}} \\

\bottomrule
\end{tabular}
} % End resizebox
\end{table}

\begin{figure}[t]
    \centering
    \includegraphics[width=0.95\linewidth]{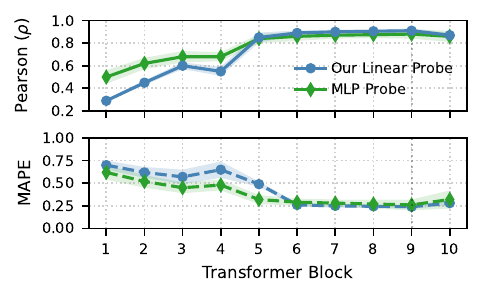}
    % \vspace{-0.2cm}
    \caption{\textbf{Probe Analysis per Block} Layer-wise performance of Non-Invasive (\textit{PhyIP}, Blue) vs. Non-Linear (MLP, Green) probes. While early layers MLP perform better than linear, the representation spontaneously \textit{linearizes} in deep layers, peaking at Block 9 ($\rho \approx 0.91$).}
    \label{fig:layer_emergence}
\end{figure}

We selected \texttt{blocks.9} via a systematic analysis per block on the 50 OOD test set (\Cref{fig:layer_emergence}). While physical information is initially entangled, requiring the non-linear MLP ($d \to 254 \to 1$) probe to extract it, the representation 'linearizes' with depth, reaching maximal disentanglement at \texttt{blocks.9} ($\rho \approx 0.91$ and MAPE $\approx 0.25$).

% --- INSERT IN SECTION 5: AFTER FIGURE 4 ---

\textbf{Mechanistic Analysis:} \Cref{fig:combined_failure} visualizes the mechanism of this failure. Using 2D PaCMAP projections of activation manifolds (methodology in \Cref{app:manifold_viz}), we observe that the SSL model ($m_\theta$) encodes a gradient of ground-truth force magnitudes. In contrast, the fine-tuned model ($m_{\theta'}$) exhibits a modified manifold, confirming the modification of geometric structure. This collapse is driven by parameter shifts in the decoder's attention and MLP layers (\Cref{app:paramchange}).  To quantify this, we analyze layer-wise activations $h^{(l)} \in \mathbb{R}^{B \times T \times d}$ for OOD test trajectories $\mathcal{X}$, where $B$ is batch size, $T$ sequence length, and $d=768$.

\begin{figure}[h]
    \centering
    \includegraphics[width=\linewidth]{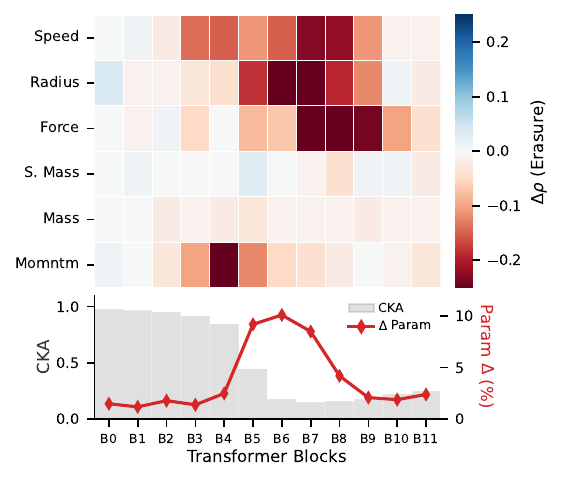}
    % \caption{\textbf{Invasiveness Mechanism.} Layer-wise analysis comparing representational similarity (CKA, blue bars) and relative parameter change (red line). While early layers remain stable, a spike in parameter updates at blocks 5--7 triggers a sharp collapse in representational similarity (CKA $< 0.2$) in the deeper layers.}
\caption{\textbf{Mechanistic Origins of Erasure.} \textbf{Top:} Heatmap of linear decodability change ($\Delta \rho$). Fine-tuning selectively erases dynamic variables (Speed, Radius) while preserving static one (Mass). \textbf{Bottom:} This collapse is driven by a parameter change causing a drop in representational similarity (CKA).}
\label{fig:mechanistic_combined}
    \label{fig:param_change}
\end{figure}

We measure \textit{Parameter Modifications}, quantified by the relative Frobenius norm of the weight change for each block $l$ : $\delta^{(l)} = \frac{\|\theta'^{(l)} - \theta^{(l)}\|_F}{\|\theta^{(l)}\|_F}$ as~\citet{guo2020broader} shows. Second, we assess \textit{Representational Drift} using Linear Centered Kernel Alignment (CKA)~\citep{kornblith2019similarity}, which computes the $s_{\text{CKA}}^{(l)}$. Finally, we identify the specific physical concepts erased by this drift.  We isolate the subset of neurons $\mathcal{S}^{(l)}$ exhibiting the top 20\% of parameter modifications $\delta_j = \|\mathbf{w}'_j - \mathbf{w}_j\|_2$ within the MLP projection layer. We then compute the shift in the maximum Pearson correlation $\Delta \rho_k$ between the flattened activation history $\mathbf{h}_j \in \mathbb{R}^N$ and ground-truth physical vectors $\boldsymbol{\phi}_k \in \mathbb{R}^N$:
\begin{equation}
            \Delta \rho_k = \max_{j \in \mathcal{S}^{(l)}} \left| \text{corr}(\mathbf{h}'_j, \boldsymbol{\phi}_k) \right| - \max_{j \in \mathcal{S}^{(l)}} \left| \text{corr}(\mathbf{h}_j, \boldsymbol{\phi}_k) \right|
\end{equation}
Applying these metrics reveals a distinct chain of corruption (\Cref{fig:mechanistic_combined}). Structural modifications are highly localized, while the global average is low, the attention and MLP layers of deep blocks (B5--B10) undergo significant shifts ($\delta^{(l)} \approx 0.10$). This structural modification directly drives functional collapse: while early layers remain stable ($s_{\text{CKA}} \approx 1.0$), the representations in these modified deep blocks diverge successfully ($s_{\text{CKA}} < 0.2$).

The results in \Cref{fig:mechanistic_combined} show a targeted erasure of dynamic invariants: encoding for \textit{Speed} and \textit{Radius} drops significantly ($\Delta \rho \approx -0.15$), while static variables like \textit{Mass} remain untouched. This confirms that fine-tuning specifically modifies parameters responsible for Hamiltonian dynamics ($ \mathcal{K} \propto v^2, 1/r$) to minimize error on the narrow distribution.

% % --- Figure 3: Physical Erasure ---
% \begin{figure}[h]
%     \centering
%     \includegraphics[width=0.8\linewidth]{Figures/mechanistic_delta_heatmap_reproduced (5) (1).pdf}
%     \caption{\textbf{Physical Erasure:} Loss of Speed and Radius information. The corruption is specific: modified neurons lose their ability to encode dynamic invariants (Speed, Radius, Force), while static concepts (Mass) are preserved.}
%     \label{fig:mechanistic_erasure}
% \end{figure}

\textbf{Fine-tuning Dataset Analysis:}
\label{sec:finetuning_data_dist}
\begin{figure}[h]
        \centering
        \includegraphics[width=0.9\linewidth]{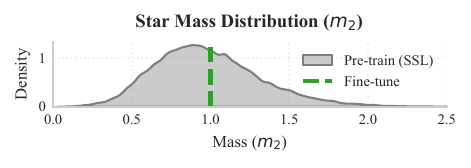}
        
    \caption{\textbf{Narrow Data.} Comparison of Star Mass ($m_2$) distributions. While the SSL pre-training data (Grey) covers a diverse physical range, the fine-tuning dataset (Green dashed) is a single point mass at $m_2 = 1.0$.}
    \label{fig:data_dist}
\end{figure}
The effectiveness of fine-tuning is heavily reliant on the dataset representativeness used for adaptation~\citep{kumar2022finetuning}. By replicating the  dataset as in \citep{vafa2025foundationmodelfoundusing}, \Cref{fig:data_dist} shows that the data for the Star Mass ($m_2$) is a single  spike at 1.0 (Green). These discrepancies indicate that the fine-tuning dataset does not provide sufficient diversity to maintain the model's general understanding of physics. This lack of support directly drives the optimizer to erase the now-unnecessary dynamic quantities (speed, radius) in favor of static shortcuts suitable for the narrow distribution (See to~\Cref{app:ftdataan} for \textit{Force Vectors} and \textit{Force Magnitude} distribution analysis).

\begin{figure}[t]
    \centering
    \includegraphics[width=\linewidth]{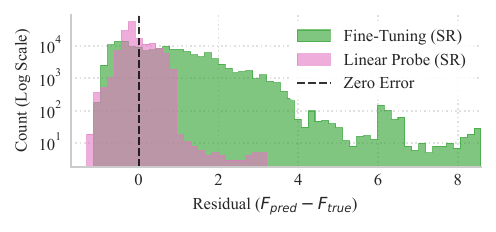}
    % \vspace{-0.2cm}
    \caption{\textbf{Symbolic Validation.} Distribution of prediction errors on OOD data. Our non-invasive probe (Blue) achieves high precision compared to the erratic invasive baseline (Green).}
    \label{fig:residual_histogram}
\end{figure}

\textbf{Symbolic Validation of Discovered Physics:}
We apply SR to distill the non-intrusive probe and IBP output into interpretable formulas as in \Cref{fig:methodology}. When evaluated them as a physical model on the 50 OOD orbital test sets (force vector task). \Cref{fig:residual_histogram} shows non-invasive probe formula tracks the ground truth with high precision, whereas the formula extracted from the IBP remains erratic. The truth is the Newton Law~\cite{newton1687}:
\(
    F \propto \frac{m_1 m_2}{r^2}
\)
The formula discovered by the IBP~\cite{vafa2025foundationmodelfoundusing} is dominated by artifacts:
\(F \propto \underbrace{\left[\sin\left(\tfrac{1}{\sin(r-0.2)}\right) + 1.5\right]}_{\text{Hallucinated Artifacts}} \cdot \underbrace{\frac{1}{r^{-1} + m_2}}_{\text{Distorted Decay}}\).

In contrast, the Non-Invasive Probe recovers the signal:
\(
    \textcolor{StageOneBlue}{F \approx \underbrace{P(r, m_2)}_{\text{Residual Noise}} + \underbrace{\mathbf{\frac{1}{r^2}}}_{\text{Recovered Law}}}
\).

Although the non-invasive probe formula is still an approximation, it discovers a distinct additive term where the residual $P(r, m_2)$ vanishes, recovering the inverse-square law $F \approx \mathbf{1/r^2}$. The full list of discovered formulas is available in Appendix \ref{app:symbolic_comparison} (\Cref{tab:symbolic_comparison_all}).

\textbf{Mechanistic Confirmation.} The successful recovery of the $\mathbf{1/r^2}$ term provides functional validation for the mechanistic analysis in \Cref{sec:finetuning}. As visualized in the erasure heatmap (\Cref{fig:mechanistic_combined}), the neurons encoding Radius ($r$) were preserved in the SSL model but \textit{specifically erased} during fine-tuning. Because the fine-tuned model lost the internal representation of $r$, it physically \textit{could not} express the correct inverse-square law, defaulting to heuristics.

\begin{takeawaypink}[title=Takeaway 3: Erasure of Encoded Dynamics]
\textit{
Invasive probes do not merely measure inductive bias—they overwrite it: fine-tuning induces representational drift that selectively \textbf{erases dynamic state variables} (e.g., speed, radius) to fit narrow downstream data. Reliable world-model evaluation must preserve the backbone.}
\end{takeawaypink}

\section{Conclusion}
AI for scientific discovery has reached a critical stage~\citep{wang2023scientific}. As models scale, the challenge shifts from training to correctly interpreting the latent knowledge they have internalized~\citep{NEURIPS2024f6a6317, mencattini2026exploratorycausalinferencesaence}.  Distinguishing whether neural dynamics models internalize physical laws as world models~\cite{NEURIPS2024f6a6317} or merely rely on statistical shortcuts~\cite{geirhos2020shortcut} is computationally difficult, as standard invasive protocols often act as interventions that distort the underlying representation.  To address this, we introduced the non-invasive \textit{PhyIP} framework to evaluate the intrinsic physics of Self-Supervised Learning (SSL) models without inducing feature distortion~\cite{kumar2022finetuning}.

Empirically, our non-invasive approach reveals physical structures that standard invasive methods miss. On complex benchmarks from ``TheWell''~\cite{howard2023well}, we precisely recovered the internal energy law ($E \approx 1.5P$) in radiative turbulence~\cite{stachenfeld2021learned} and identified emergent correction terms for convective kinetic energy ($\sim \rho v_r^2$) in stellar simulations ($\rho > 0.90$)~\cite{Goldberg_2022}. Furthermore, by replicating orbital mechanics experiments~\citep{vafa2025foundationmodelfoundusing}, we successfully recovered the inverse-square law with high fidelity ($\rho \approx 0.91$), whereas invasive adaptation probes reported near-zero correlation ($\rho \approx 0.05$).

Conversely, we show that invasive probes—including nonlinear MLP probes~\cite{belinkov2022probing}, Last-Layer Fine-Tuning (LLFT)~\cite{kirichenko2023last}, and Inductive Bias Probes (IBP)~\cite{vafa2025foundationmodelfoundusing} can act as destructive interventions. Rather than passively measuring latent knowledge, they overwrite the representation~\citep{kumar2022finetuning}. Our mechanistic analysis confirms that these optimizers systematically suppress complex time-varying features (e.g., speed, radius) to exploit simple constant identifiers (e.g., mass), effectively ``hallucinating'' competence or erasing physical laws to fit narrow data~\citep{mukhoti2024finetuning, geirhos2020shortcut}.

% Finally, our work generalizes the architectural constraints proposed by ~\citep{liu2026keplernewtoninductivebiases}. While they argue that specific inductive biases—such as strict temporal locality via a context window must be architecturally forced into the model to learn Newtonian dynamics, we demonstrate that PhyIP can decode these dynamics linearly from the SSL models. The ``Newtonian'' world model is not missing; it is latent. By using \textit{PhyIP}, we impose a simplicity constraint analytically rather than architecturally, allowing us to recover fundamental laws like $F \propto 1/r^2$ without the need for restrictive retraining.

% Finally, we offer a parsimonious alternative to the 'Keplerian vs. Newtonian' dichotomy presented by~\citep{liu2026keplernewtoninductivebiases}. ~\citep{liu2026keplernewtoninductivebiases} force the emergence of Newtonian dynamics by restricting the attention window, physically preventing the model from memorizing historical trajectories. In contrast, we show that this inductive bias does not need to be hard-coded. SSL can naturally encode the undeling law ($F \propto 1/r^2$) linearly.

In concurrent work, \citet{liu2026keplernewtoninductivebiases} identify key inductive biases; specifically spatial smoothness (continuous regression), stability (noise injection), and temporal locality (context restriction); that enable Transformers to learn Newtonian physics with perfect fidelity ($R^2 \approx 1$). While they demonstrate that explicitly enforcing these constraints guarantees the acquisition of exact physical models, our work offers a complementary perspective focused on evaluation. We find that even without these additional inductive biases, standard SSL approximately encodes physical dynamics into linear subspaces. Although these latent representations may not always reach the perfect precision of constrained models, their linear extractability confirms that the core physical laws are already emerging. This validates the promise of general-purpose foundation models \citep{bommasani2022opportunitiesrisksfoundationmodels}: that broad physical understanding can emerge implicitly from data scale and diversity.

% In related work, \citep{liu2026keplernewtoninductivebiases} explore the effect of context window length, demonstrating that retraining with specific constraints can effectively induce Newtonian dynamics. However, our results suggest that such hard-coded constraints may not be strictly necessary for the emergence of physical laws. We find evidence that standard SSL, is capable of naturally encoding laws like $F \propto 1/r^2$ into linear subspaces. This indicates that the necessary inductive biases can emerge from the SSL task itself, rather than requiring explicit architectural enforcement.

We hope this work encourages the adoption of neural models as \textit{fixed measurement instruments}~\cite{10.1111/rssb.12167} and the use of non-invasive protocols to distinguish true machine learning failures from artifacts of adaptation. Ultimately, our findings suggest that \textit{Scientific AI requires not just better models, but better instruments to measure them.}

\textbf{Limitations and Future Work:} We identify three primary constraints. Linear probes prevent the probe from solving the physics independently but limit performance on highly non-linear entangled representations, where physical quantities may be encoded in more complex, non-linear geometries. Future research should investigate subspace-constrained or weight-preserving adaptation protocols. These methods would aim to acquire new task-specific capabilities while strictly protecting the linear physical invariants—such as conservation laws—already internalized by the model.

\nocite{langley00}

\bibliography{example_paper}
\bibliographystyle{icml2026}

%%%%%%%%%%%%%%%%%%%%%%%%%%%%%%%%%%%%%%%%%%%%%%%%%%%%%%%%%%%%%%%%%%%%%%%%%%%%%%%
%%%%%%%%%%%%%%%%%%%%%%%%%%%%%%%%%%%%%%%%%%%%%%%%%%%%%%%%%%%%%%%%%%%%%%%%%%%%%%%
% APPENDIX
%%%%%%%%%%%%%%%%%%%%%%%%%%%%%%%%%%%%%%%%%%%%%%%%%%%%%%%%%%%%%%%%%%%%%%%%%%%%%%%
%%%%%%%%%%%%%%%%%%%%%%%%%%%%%%%%%%%%%%%%%%%%%%%%%%%%%%%%%%%%%%%%%%%%%%%%%%%%%%%
\newpage
\appendix
\onecolumn
\section*{Appendix Contents}
\vspace{0.5em}
\noindent\rule{\linewidth}{0.8pt} % A thicker top rule
\vspace{1.2em}

% Define custom commands for the appendix ToC entries (if not already defined)
\newcommand{\tocsection}[2]{%
    \noindent\hyperref[#1]{\textbf{\ref*{#1}}\quad#2}\hfill\pageref{#1}\par\vspace{3pt}}
\newcommand{\tocsubsection}[2]{%
    \noindent\hspace{2em}\hyperref[#1]{\ref*{#1}\quad#2}\ \textcolor{black!50}{\dotfill}\ \pageref{#1}\par}

% --- The Table of Contents List for YOUR paper ---
% \tocsection{app:appendix_A}{Appendix A} % Assuming you have a general % A. Related Work
\tocsection{app:related_work}{Additional Related Work} 

% B. Formal Derivation (Previously Missing)
\tocsection{app:proof}{Formal Derivation of Equation (2)} 

% C. Fluid Dynamics (Moved to correct alphabetical position)
\tocsection{app:fluid_dynamics_results}{Fluid Dynamics Experiment Setting} 

% D. Manifold Visualization
\tocsection{app:manifold_viz}{Manifold Visualization Methodology} 

% E. Solar System
\tocsection{app:solar_system_analysis}{Analysis of Solar System Replication} 

% F. Parameter Change (The ``Param Change'' section)
\tocsection{app:paramchange}{Parameter Change Analysis} 

% G. Data Distribution (Previously Missing)
\tocsection{app:ftdataan}{Full Fine-tuning Data Distribution Analysis} 

% H. Symbolic Formulas
\tocsection{app:symbolic_comparison}{Symbolic Formula Comparison}
\vspace{1em}
\noindent\rule{\linewidth}{0.4pt} % A thinner bottom rule
\vspace{2em}

\section{Additional Related Work}
\label{app:related_work}

We situate our work at the intersection of mechanistic interpretability and the dynamics of transfer learning for AI-driven scientific discovery.

\paragraph{AI for Physics.}
The discovery of physical laws from data typically relies on two paradigms: enforcing laws via architectural priors, such as Hamiltonian/Lagrangian Neural Networks (HNNs/LNNs) \citep{greydanus2019hamiltonian, cranmer2020lagrangian, karniadakis2021physicsinformed}, or post-hoc Symbolic Regression (SR) \citep{schmidt2009distilling}. While SR methods attempt to learn symbolic expressions directly from high-dimensional inputs \citep{biggio2021neural, kamienny2022endtoend}, they often struggle with the curse of dimensionality.
However, recent findings suggest that standard SSL suffices to identify dynamics without physics-specific constraints \citep{chen2022automated, reizinger2025identifiability}. \citet{internò2025aigeneratedvideodetectionperceptual} observe that physically consistent dynamics from natural video emerge as linear ``straight'' trajectories in pre-trained latent spaces,  whereas AI-generated video creates curved latent trajectories due to physical artifact implausibility.

% \paragraph{Inductive Biases and Causal Discovery.}
% Debate persists on whether foundation models capture physical laws. \citet{vafa2025foundationmodelfoundusing} argue generic Transformers fail, relying on statistical shortcuts. \citet{liu2026keplernewtoninductivebiases} attribute this to missing constraints, distinguishing between ``Keplerian'' (curve-fitting) and ``Newtonian'' (causal) models, where the latter requires enforced biases like temporal locality. Conversely, \citet{spies2025transformersusecausalworld} detect latent causal structures in standard models. Unifying these via the Linear Representation Hypothesis \citep{nanda2023emergentlinearrepresentationsworld}, we demonstrate that Newtonian dynamics are already latent in unconstrained SSL. Our non-invasive probe \textit{PhyIP} acts as a ``simplicity filter,'' decoding these linear dynamics while ignoring complex Keplerian artifacts.

\paragraph{Inductive Biases and Causal Discovery.}
The mechanism by which generic foundation models capture physical laws remains a subject of intense debate. \citet{vafa2025foundationmodelfoundusing} utilized ``Inductive Bias Probes'' to audit models for physical compliance, concluding that standard Transformers achieve high predictive accuracy but fail to internalize fundamental forces. Responding to this, \citet{liu2026keplernewtoninductivebiases} argue that this failure stems from a lack of architectural constraints; they distinguish between ``Keplerian'' world models (curve-fitting) and ``Newtonian'' models (causal forces), demonstrating that the latter only emerge when specific biases are enforced.
However, recent mechanistic interpretability studies suggest that causal structures may emerge naturally without such constraints~\cite{nanda2023emergentlinearrepresentationsworld}. \citet{spies2025transformersusecausalworld} identified latent ``World Models'' in maze-solving Transformers, while \citet{mencattini2026exploratorycausalinferencesaence} demonstrated that causal effects can be explicitly recovered from frozen foundation models using Sparse Autoencoders~\cite{klindt2025superpositionsparsecodesinterpretable}, effectively disentangling the ``treatment'' variables from unstructured representations.

\paragraph{Self-Supervised Learning and Emergent World Models.}

The capability of next-token prediction to induce compact belief states remains a subject of active debate. \citet{teoh2025nextlatentpredictiontransformerslearn} argue that in generic discrete domains, this objective is theoretically insufficient without auxiliary losses. This view is challenged by empirical findings in game playing \citep{nanda2023emergentlinearrepresentationsworld} and mechanistic analysis \citep{nanda2023progressmeasuresgrokkingmechanistic}, which demonstrate that Transformers develop linear representations of ``world models''—solely from the predictive objective.
From a theoretical standpoint, work on identifiability in SSL \citep{pmlrzimmermann21a, kgelgen2021selfsupervised} suggests that it can provably recover ground-truth latent factors given sufficient data diversity.
In the context of continuous physics dynamics, we align with the view that residual networks function as discretizations of Ordinary Differential Equations~\citep{Haber_2018, chen2018neural}.

\section{Formal Derivation of Equation (2)}
\label{app:proof}
\begin{figure}[ht]
\centering
\begin{tikzpicture}[>=stealth, scale=1.2, font=\small\sffamily]

    % --- Definitions & Coordinates ---
    \coordinate (O) at (0,0);
    
    % 1. The True Physics Vector (Delta x)
    \coordinate (dx) at (4, 0.5);
    
    % 2. The Model Update Vector (J h_t) - Slight error from Truth
    \coordinate (Jh) at (3.8, 1.2);
    
    % 3. The Local Gradient (Nabla Phi) - Perpendicular to local level set
    \coordinate (Grad) at (1, 2.5);
    
    % 4. The Mean Gradient (Mu_Phi) - The Fixed Probe's direction
    % Deviates from Local Gradient due to Curvature K
    \coordinate (MeanGrad) at (2.2, 2.2);

    % --- Visual Elements ---

    % A. The Manifold / Level Sets (Background)
    \begin{scope}
        \clip (-1,-1) rectangle (6, 4);
        % Curved lines representing the non-linear physical functional Phi
        \draw[gray!20, thin] (-1, -0.5) to[out=20, in=200] (6, 1.5);
        \draw[gray!30, thin] (-1, 0.5) to[out=20, in=200] (6, 2.5);
        \draw[gray!20, thin] (-1, 1.5) to[out=20, in=200] (6, 3.5);
    \end{scope}

    % B. Vectors
    
    % True Update
    \draw[->, very thick, black] (O) -- (dx) node[midway, below] {$\Delta x_t$ (True)};
    
    % Model Update (Linearized Decoder)
    \draw[->, very thick, StageOneBlue] (O) -- (Jh) node[midway, above left, text=StageOneBlue] {$J h_t$ (Model)};
    
    % Local Gradient (Linearized Physics)
    \draw[->, thick, StageThreePink] (O) -- (Grad) node[above, text=StageThreePink] {$\nabla \Phi(x_t)$};
    
    % Mean Gradient (Fixed Probe)
    \draw[->, thick, StageTwoOrange] (O) -- (MeanGrad) node[above right, text=StageTwoOrange] {$\mu_\Phi$ (Probe)};

    % --- ERROR VISUALIZATION ---

    % 1. Modeling Error (Gap between Model and Truth)
    \draw[dashed, color=StageOneBlue!80] (dx) -- (Jh) 
        node[midway, right, font=\footnotesize, align=left] {Modeling\\Error ($\epsilon$)};

    % 2. Curvature Error (Angle between Local and Mean Gradient)
    % Draw an arc to show the angle deviation
    \draw[<->, thick, color=physred] (0.6, 1.5) to[bend left=20] (1.3, 1.3);
    \node[text=physred, font=\footnotesize] at (1.0, 1.8) {Curvature ($K_\Phi$)};

    % --- Projections (Dot Products) ---
    % Dotted lines to show the projections onto the gradients
    % This represents the scalar value the probe predicts vs reality
    
    % Truth Projection onto Local Gradient
    \draw[dotted, gray] (dx) -- ($(O)!(dx)!(Grad)$);
    
    % Probe Projection onto Mean Gradient
    \draw[dotted, gray] (Jh) -- ($(O)!(Jh)!(MeanGrad)$);

\end{tikzpicture}
\caption{\textbf{Geometry of the Linear Probe Error Bound.} The total error stems from two mismatches: 
\textbf{1) Modeling Error:} The deviation of the model's linearized update ($J h_t$) from the true physical update ($\Delta x_t$).
\textbf{2) Curvature Error:} The angular deviation between the local physical gradient ($\nabla \Phi_t$) and the global mean gradient ($\mu_\Phi$) used by the fixed linear probe. High curvature ($K_\Phi$) increases this angle.}
\label{fig:derivation_geometry}
\end{figure}
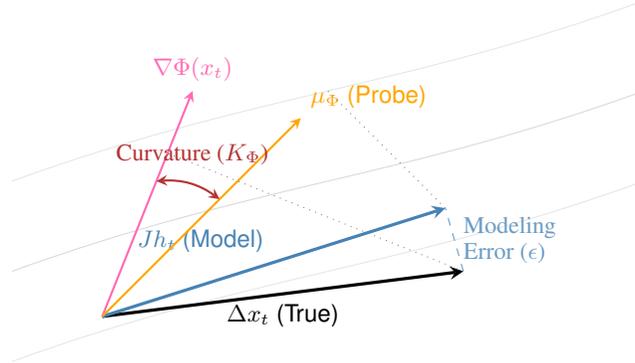

We derive the error bound for the linear probe, explicitly handling the approximation errors induced by the non-linearity of the physical functional and the neural decoder.

\subsection{Proof Strategy: The Double Linearization}
Our goal is to bound the error of a linear probe $P_{W^*}$ mapping the latent representation $h_t$ directly to the physical quantity update $\Delta s_t$. To do this, we decompose the true non-linear transition into two approximating linear steps:

\begin{enumerate}
    \item \textbf{Physics Linearization:} We approximate the curved physical law $\Delta s_t = \Phi(x_{t+1}) - \Phi(x_t)$ via its local gradient $\nabla \Phi(x_t)$. The error in this step depends on the \textit{Physical Curvature} $K_\Phi$.
    \item \textbf{Model Linearization:} We approximate the non-linear residual decoder $g(h_t)$ via its Jacobian $J$. The error in this step is bounded by the \textit{SSL Prediction Error} $\epsilon$.
\end{enumerate}

The total probe error is derived by bounding the mismatch between the fixed linear probe $W^*$ (which must average over the state space) and these local linearizations.

\subsection{Setup and Definitions}
Let the state space be $\mathcal{X} \subset \mathbb{R}^n$. The physical functional is $\Phi: \mathcal{X} \to \mathbb{R}^k$, such that $s_t = \Phi(x_t)$. The target update is $\Delta s_t = s_{t+1} - s_t$.
The model predicts $\hat{x}_{t+1} = x_t + g(h_t)$, where $g(\mathbf{0}) = \mathbf{0}$.

\begin{itemize}
    \item \textbf{Physics Dynamics:} $x_{t+1} = x_t + \Delta x_t$, where $\Delta x_t \approx \Delta t \cdot f(x_t)$.
    \item \textbf{SSL Objective:} $\mathbb{E}[\|\hat{x}_{t+1} - x_{t+1}\|^2] \le \epsilon$.
    \item \textbf{Linear Probe:} $P_{W^*}(h_t) = W^* h_t$, where $W^* \in \mathbb{R}^{k \times d}$.
\end{itemize}

\subsection{Step-by-Step Derivation}

\paragraph{Step 1: Linearizing the Physical Law ($\Phi$).}
Assuming $\Phi$ is $C^2$-smooth, we expand the physical update around the current state $x_t$:
\begin{equation}
    \Delta s_t = \nabla \Phi(x_t)^\top \Delta x_t + R_\Phi(x_t, \Delta t)
\end{equation}
where the remainder is bounded by the curvature constant $K_\Phi$ (the Lipschitz constant of $\nabla \Phi$): 
\begin{equation}
    \|R_\Phi\| \le \frac{1}{2} K_\Phi \|\Delta x_t\|^2 \in \mathcal{O}(\Delta t^2)
\end{equation}

\paragraph{Step 2: Linearizing the Model Decoder ($g$).}
Since the decoder is origin-preserving, we linearize around the null latent $h_t=\mathbf{0}$:
\begin{equation}
    g(h_t) = J h_t + R_g(h_t)
\end{equation}
where $J = \nabla g(\mathbf{0}) \in \mathbb{R}^{n \times d}$ is the Jacobian. The SSL error constraint $\mathbb{E}[\|g(h_t) - \Delta x_t\|^2] \le \epsilon$ implies that the linear term $J h_t$ approximates the true physical update $\Delta x_t$ up to the training error and higher-order terms. Specifically, $\mathbb{E}[\|J h_t - \Delta x_t\|^2] \approx \epsilon$.

\paragraph{Step 3: Defining the Optimal Fixed Probe.}
A linear probe $W^*$ must correspond to a single, time-independent matrix. The optimal choice is the projection of the \textit{expected global gradient} onto the decoder's tangent space:
\begin{equation}
    W^* = \mu_\Phi^\top J
\end{equation}
where $\mu_\Phi = \mathbb{E}_{x}[\nabla \Phi(x)] \in \mathbb{R}^{n \times k}$ is the mean gradient of the functional over the state distribution.

\paragraph{Step 4: Error Decomposition.}
We analyze the squared error of the probe prediction against the true update:
\begin{equation}
    \mathcal{L}_{\text{probe}} = \mathbb{E}[ \| W^* h_t - \Delta s_t \|^2 ]
\end{equation}
Substituting the linearizations from Steps 1 and 2, and adding/subtracting the term $\nabla \Phi(x_t)^\top J h_t$ (the local linear approximation):

\begin{align}
    \| W^* h_t - \Delta s_t \| &= \| \mu_\Phi^\top J h_t - (\nabla \Phi(x_t)^\top \Delta x_t + R_\Phi) \| \notag \\
        &= \| \underbrace{(\mu_\Phi - \nabla \Phi(x_t))^\top J h_t}_{\text{Term A: Curvature Mismatch}} + \underbrace{\nabla \Phi(x_t)^\top (J h_t - \Delta x_t)}_{\text{Term B: Modeling Error}} - R_\Phi \|
\end{align}
Using the inequality $(a+b+c)^2 \le 3(a^2 + b^2 + c^2)$, we bound the expectation:

\textbf{Analyzing Term A (Curvature Mismatch):}
This term measures the error of using the \textit{average} gradient $\mu_\Phi$ instead of the \textit{local} gradient $\nabla \Phi(x_t)$. Applying Cauchy-Schwarz:
\begin{equation}
    \mathbb{E}[\text{Term A}^2] \le \mathbb{E}[ \| \mu_\Phi - \nabla \Phi(x_t) \|^2 \| J h_t \|^2 ]
\end{equation}
We bound the step size $\|J h_t\|^2 \le C_{\text{step}}$. The remaining term is strictly the variance of the gradient, $\operatorname{Var}(\nabla \Phi(x))$. Using the Lipschitz property of the gradient (Curvature $K_\Phi$):
\begin{equation}
    \operatorname{Var}(\nabla \Phi(x)) \le K_\Phi^2 \cdot \operatorname{Var}(x)
\end{equation}
Thus, $\mathbb{E}[\text{Term A}^2] \le C_{\text{step}} K_\Phi^2 \operatorname{Var}(x)$.

\textbf{Analyzing Term B (Modeling Error):}
This term measures the failure of the model to produce the correct state update.
\begin{equation}
    \mathbb{E}[\text{Term B}^2] \le \sup_x \|\nabla \Phi(x)\|^2 \cdot \mathbb{E}[\|J h_t - \Delta x_t\|^2] \le C_{\text{grad}} \cdot \epsilon
\end{equation}

\paragraph{Final Bound.}
Combining terms, we obtain the final inequality:
\begin{equation}
    \mathbb{E} \left[ \| P_{W^*} h_t - \Delta s_t \|^2 \right] \le C_1 \cdot \epsilon + C_2 \left[ K_{\Phi}^2 \cdot \operatorname{Var}(x) \right] + \mathcal{O}(\Delta t^4)
\end{equation}

\section{Fluid Dynamics Experiment Settings}
\label{app:fluid_dynamics_results}

To validate the generality of our non-invasive probing framework, we applied it to three complex, high-dimensional fluid dynamics simulations from \textit{TheWell} benchmark \citep{howard2023well}. We tested multiple neural simulator architectures—U-Net \citep{ronneberger2015u}, UNetConvNext \citep{unet2020}, FNO \citep{li2021fourier}, and TFNO \citep{li2021fourierneuraloperatorparametric}—all trained on a self-supervised next-step prediction task.

Our objective was to determine if these models implicitly learned the conservation of total internal energy ($E_{\text{int}}$) purely from observing state transitions. We extracted frozen activations $h(t)$ from the bottleneck (U-Nets) or the final spectral block (FNOs) and trained a linear probe to predict $E_{\text{int}}(t+1)$.

\subsection{2D Turbulent Radiative Layer (TRL-2D)}
This simulation models a 2D slice of a stellar atmosphere or accretion disk, governed by compressible magnetohydrodynamics (MHD) with radiative transfer. It captures the interplay between magnetic turbulence and radiative cooling.
\noindent
\begin{minipage}[c]{0.65\linewidth}
    \paragraph{Governing Equations.}
    The system evolves according to:
    \begin{align}
        \frac{\partial \rho}{\partial t} + \nabla \cdot (\rho v) &= 0 \\
        \frac{\partial (\rho v)}{\partial t} + \nabla \cdot (\rho v v + P) &= 0 \\
        \frac{\partial E}{\partial t} + \nabla \cdot ((E + P) v) &= -\frac{E}{t_{\text{cool}}}
    \end{align}
    where the internal energy is defined by the ideal gas law: $E = P/(\gamma - 1)$ with $\gamma = 5/3$.
\end{minipage}%
\hfill
\begin{minipage}[c]{0.32\linewidth}
    \centering
    % FIX: Used \fbox instead of 'frame' option
    \setlength{\fboxsep}{0pt}
    \fbox{\includegraphics[width=\linewidth]{Figures/Capture11.PNG}}
    \captionof{figure}{TRL-2D Simulation \citep{stachenfeld2021learned}}
\end{minipage}

% \paragraph{Governing Equations.}
% The system evolves according to:
% \begin{align}
%     \frac{\partial \rho}{\partial t} + \nabla \cdot (\rho \mathbf{v}) &= 0 \\
%     \frac{\partial (\rho \mathbf{v})}{\partial t} + \nabla \cdot (\rho \mathbf{v} \mathbf{v} + P) &= 0 \\
%     \frac{\partial E}{\partial t} + \nabla \cdot ((E + P) \mathbf{v}) &= -\frac{E}{t_{\text{cool}}}
% \end{align}
% where the internal energy is defined by the ideal gas law: $E = P/(\gamma - 1)$ with $\gamma = 5/3$.

\paragraph{Task \& Probe Configuration.}
\begin{itemize}[leftmargin=*, noitemsep]
    \item \textbf{SSL Input:} $\{\rho, v_x, v_y, P\}$ at time $t$. Resolution: $128 \times 128$.
    \item \textbf{Probe Target:} The total internal energy $E_{\text{int}} = \int_{\Omega} \frac{P}{\gamma-1} dV$.
    % \item \textbf{Challenge:} Since $E \propto P$, the mapping is linear ($K_\Phi = 0$). However, the \textit{dynamics} $\partial E / \partial t$ involve non-linear advection. Success confirms the model has disentangled the linear static law from the non-linear transport.
\end{itemize}

\subsection{3D Red Supergiant Convective Envelope (RSG-3D)}
This dataset simulates the outer convective envelope of a red supergiant star, governed by compressible hydrodynamics with radiative transfer. It features strong convective upflows and buoyancy-driven turbulence.
\noindent
\begin{minipage}[c]{0.65\linewidth}
    \paragraph{Governing Equations.}
    \begin{align}
        \frac{d\rho}{dt} &= -\rho \nabla \cdot V \\
        \frac{d^2\mathbf{r}}{dt^2} &= -\frac{\nabla P}{\rho} + \mathbf{a}_{\text{visc}} - \nabla \Phi_{\text{grav}} \\
        \frac{du}{dt} &= -\frac{P}{\rho} \nabla \cdot V + \frac{\Gamma - \Lambda}{\rho}
    \end{align}
    Here, $u$ is specific internal energy, $\Phi_{\text{grav}}$ is gravitational potential, and $\Gamma, \Lambda$ represent radiative heating/cooling.
\end{minipage}%
\hfill
\begin{minipage}[c]{0.32\linewidth}
    \centering
    % FIX: Used \fbox instead of 'frame' option
    \setlength{\fboxsep}{0pt}
    \fbox{\includegraphics[width=\linewidth]{Figures/CaptureRd1.PNG}}
    \captionof{figure}{RSG-3D Simulation \citep{Goldberg_2022}}
\end{minipage}

\paragraph{Task \& Probe Configuration.}
\begin{itemize}[leftmargin=*, noitemsep]
    \item \textbf{SSL Input:} $\{\rho, P, v_r, v_\theta, v_\phi\}$ at time $t$. Resolution: $64 \times 64 \times 64$.
    \item \textbf{Probe Target:} Total internal energy $E_{\text{int}} = \int_{\Omega} \rho u \, dV$. Note that specific energy $u$ is not an input; the probe must implicitly derive $u$ from $P$ and $\rho$ via the equation of state.
    % \item \textbf{Insight:} As shown in Table \ref{tab:merged_fluid_results}, our probe discovers a correction term dependent on convective velocity ($v_r^2$), indicating the model accounts for kinetic energy conversion in the convection zone.
\end{itemize}

\subsection{3D Supernova Explosion (SN-3D)}
A simulation of a core-collapse supernova, involving extreme relativistic velocities, shock waves, and a simplified nuclear burning network.

\noindent
\begin{minipage}[c]{0.65\linewidth}
    \paragraph{Governing Equations.}
    The system includes the standard conservation of mass and momentum, but energy is dominated by nuclear terms:
    \begin{align}
         \frac{\partial E}{\partial t} + \nabla \cdot \dots &= -c G^0_r - \rho V \cdot \nabla \Phi \\
         \frac{\partial I}{\partial t} + c \mathbf{n} \cdot \nabla I &= S(I, \mathbf{n})
    \end{align}
    where $S$ is the source term from the nuclear burning network, creating extreme non-linearities.
\end{minipage}%
\hfill
\begin{minipage}[c]{0.32\linewidth}
    \centering
    % FIX: Used \fbox instead of 'frame' option
    \setlength{\fboxsep}{0pt}
    \fbox{\includegraphics[width=0.8\linewidth]{Figures/CaptureSN1.PNG}}
    \captionof{figure}{SN-3D Simulation \citep{hirashima2023surrogate}}
\end{minipage}

\paragraph{Task \& Probe Configuration.}
\begin{itemize}[leftmargin=*, noitemsep]
    \item \textbf{SSL Input:} $\{\rho, P_{\text{gas}}, T, v_x, v_y, v_z\}$. Resolution: $64 \times 64 \times 64$.
    \item \textbf{Probe Target:} Internal energy density $\epsilon = \rho u$, derived from a lookup table of nuclear equations of state (EOS), not a simple ideal gas law.
    % \item \textbf{Failure Mode:} The high aliasing error ($\epsilon_{\text{SSL}} > 0.3$) prevents the formation of a stable linear subspace for energy.
\end{itemize}

\subsection{Data Partitioning}
\label{app:data_split}
To ensure the validity of OOD generalization claims, we enforced a strict separation of datasets following setting descibed in \cite{howard2023well}. 
\begin{enumerate}
    \item \textbf{SSL Training Set ($D_{\text{train}}$):} Used \textit{only} for pre-training the backbone model $m_\theta$.
    \item \textbf{Probe Training Set ($D_{\text{probe}}$):} A strictly In-Distribution (ID) subset held out from $D_{\text{train}}$. The probe learns the mapping $W$ on this data. \textit{Crucially, the probe never sees OOD data during training.}
    \item \textbf{OOD Test Sets ($D_{\text{OOD}}$):} Completely distinct simulations with physical parameters (e.g., cooling rates, stellar mass) unseen in either $D_{\text{train}}$ or $D_{\text{probe}}$.
\end{enumerate}

\subsection{Implementation and Reproducibility Details}
\label{app:reproducibility}

We detail the training hyper-parameters for the SSL backbone, the non-invasive probes, and the invasive baselines in Table \ref{tab:final_specs}.

\textbf{SSL Training.} All models were trained using the AdamW optimizer with a cosine annealing schedule. Training was performed on 2$\times$ NVIDIA H100 GPUs.

\textbf{Probe Training.} The linear probes were trained using Ridge Regression (L2 regularization) on the frozen representations.

\textbf{Invasive Baselines.}
\begin{itemize}[noitemsep, topsep=0pt]
    \item \textbf{MLP Probe:} A non-linear probe composed of two dense layers ($d \to 254 \to 1$) with ReLU activation. Trained using Adam on the frozen representation.
    \item \textbf{Full Fine-tuning (IBP):} The entire backbone $m_\theta$ is unfrozen and updated to predict the physical target exactly as in \cite{vafa2025foundationmodelfoundusing}.
\end{itemize}

\begin{table}[h]
\centering
\small
\setlength{\tabcolsep}{3.5pt}
\caption{\textbf{Dataset \& Hyperparameter Specifications.} Comparison of the Newtonian Two-Body task \cite{vafa2025foundationmodelfoundusing} and The Well benchmarks~\cite{howard2023well}.}
\label{tab:final_specs}
\resizebox{0.8\linewidth}{!}{%
\begin{tabular}{l c c c c}
\toprule
\textbf{Feature} & \textbf{Two-Body (Newton)} & \textbf{TRL-2D} & \textbf{RSG-3D} & \textbf{SN-3D} \\
\midrule
Spatial Resolution & $2$ ($x,y$ coords) & $384 \times 128$ & $256 \times 128 \times 256$ & $64^3$ \\
Total Trajectories & $10 \times 10^6$ & 90 & 29 & 740 \\
\midrule
\rowcolor{gray!10} \multicolumn{5}{l}{\textit{\textbf{Data Partitioning (Train / Probe / OOD)}}} \\
SSL Train ($N_{SSL}$) & $10 \times 10^6$ & 72 & 23 & 592 \\
Probe Train ($N_{Probe}$) & 10,000 & 9 & 3 & 74 \\
OOD Test Set & 5 Galaxies & 9 Cooling Rates & 3 Phases & 27 Env. Vars \\
\midrule
\rowcolor{gray!10} \multicolumn{5}{l}{\textit{\textbf{Probe \& Adaptation Hyperparameters}}} \\
Optimizer & AdamW & AdamW & AdamW & AdamW \\
Batch Size & 64 & 32 & 8 & 8 \\
Ridge Reg. ($\alpha$) & 1.0 & 1.0 & 1.0 & 1.0 \\
MLP Hidden Dim & 254 & 254 & 254 & 254 \\
Adaptation LR & $1\text{e-}3$ & $1\text{e-}3$ & $1\text{e-}3$ & $1\text{e-}3$ \\
\bottomrule
\end{tabular}
}
\end{table}

\section{Manifold Visualization Methodology}
\label{app:manifold_viz}

The manifold visualizations in \Cref{fig:combined_failure} (Top) are generated to contrast the internal representation geometry of the pre-trained SSL model ($m_\theta$) and the fine-tuned model ($m_{\theta'}$). The process, as implemented in the provided analysis script, is as follows:

\begin{itemize}[leftmargin=*]
    \item \textit{1. Data Sampling:} We sample 50,000 data points from the validation set. For each point, we compute and store its corresponding ground-truth force magnitude ($\|\vec{F}\| = \sqrt{F_x^2 + F_y^2}$), which is used to color the final plot.

    \item \textit{2. Activation Extraction:} We process these 50,000 inputs through both the frozen pre-trained model and the fine-tuned model. Using a PyTorch forward hook, we extract the high-dimensional activation vectors (embedding dimension $d=768$) from the \textit{input} to the 10th decoder block (\texttt{'blocks.9'}). This layer is chosen to match the layer used for linear probing. This step yields two distinct sets of high-dimensional representations: $h_{ssl} \in \mathbb{R}^{50000 \times 768}$ from the SSL model and $h_{ft} \in \mathbb{R}^{50000 \times 768}$ from the fine-tuned model.

    \item \textit{3. Dimensionality Reduction:} To visualize these 768-dimensional manifolds in 2D, we apply the PaCMAP dimensionality reduction algorithm \citep{JMLR:v22:20-1061}. Crucially, we fit the PaCMAP algorithm \textit{independently} to each set of activations. This generates two 2D projections, $Z_{ssl} = \text{PaCMAP}(h_{ssl})$ and $z_{ft} = \text{PaCMAP}(h_{ft})$. While other algorithms like t-SNE and UMAP were considered, we selected PaCMAP for its superior ability to preserve both \textit{local} and \textit{global} data structure.
\end{itemize}

\section{Solar System Experiment}
\label{app:solar_system_analysis}

\begin{figure}[ht]
\centering
\includegraphics[width=0.9\linewidth]{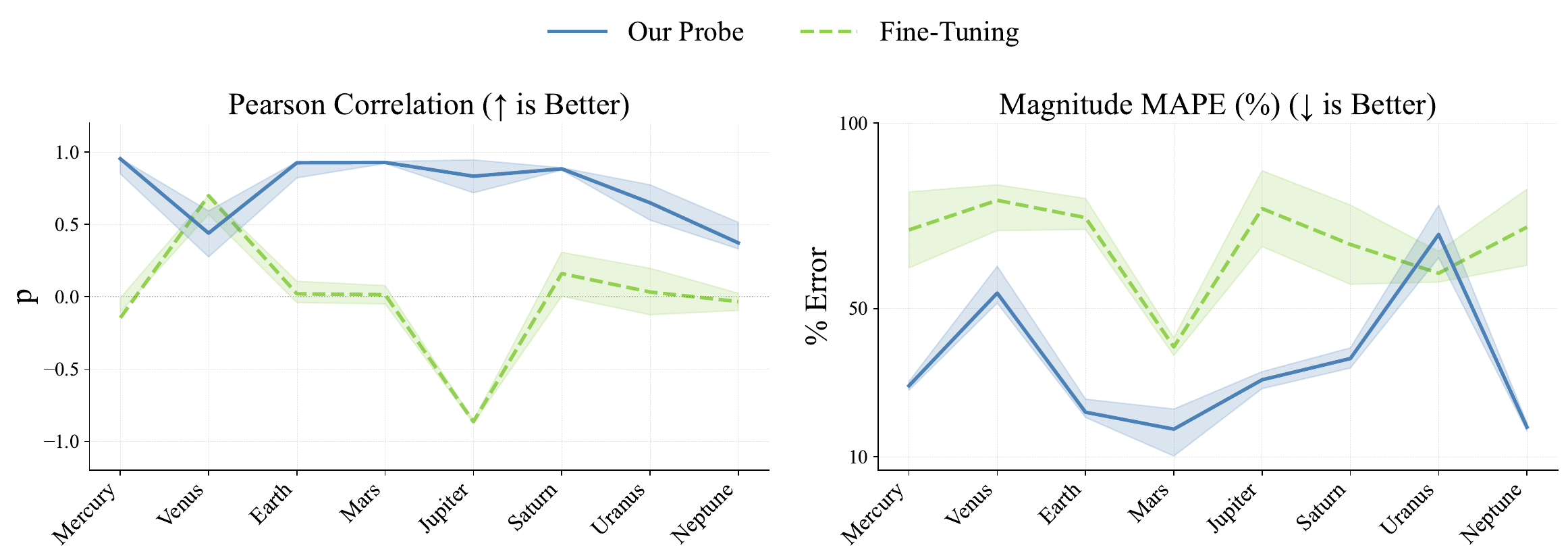}
\caption{\textbf{Solar System Generalization Analysis.} Comparison of force vector prediction performance. The invasive probe (Green) fails systematically, exhibiting high variance and negative correlations. The non-invasive probe (Blue) remains robust, though it highlights specific OOD challenges for Venus and Uranus.}
\label{fig:solar_system_comparison}
\end{figure}

We replicated the true solar system setup from \citet{vafa2025foundationmodelfoundusing} to evaluate OOD generalization on real orbital dynamics. As shown in \Cref{fig:solar_system_comparison}, our non-invasive probe (blue) consistently outperforms the invasive fine-tuning baseline (green), which exhibits erratic behavior across the planetary suite.

\paragraph{Outlier Analysis (Venus \& Uranus).}
While the non-invasive probe generally succeeds, we identify two distinct failure modes. For \textit{Venus}, the probe's Pearson correlation drops to $\rho \approx 0.4$ with a corresponding spike in error (MAPE $> 50\%$). Conversely, for \textit{Uranus}, the probe maintains high linearity ($\rho \approx 0.8$) but suffers from calibration error (MAPE $\approx 70\%$). This suggests that while the model correctly encodes the \textit{directionality} of the force at Uranus's distance, the magnitude scaling at the outer solar system boundary drifts from the training distribution.

\paragraph{Systemic Failure of Invasive Probing.}
In contrast, the failure of the invasive fine-tuning probe is not merely an issue of precision, but of fundamental physical correctness. For Jupiter, the invasive probe exhibits a strong \textit{negative} Pearson correlation ($\rho \approx -0.9$). This indicates that the fine-tuning process has inverted the vector field, effectively predicting a repulsive force rather than an attractive one. For the inner planets, the invasive probe shows near-zero correlation ($\rho \approx 0.0$), confirming the mechanistic finding in \Cref{sec:finetuning} that dynamic invariants (like the distance-force relationship) are erased during adaptation.

Our non-invasive probe maintains high performance ($\rho > 0.85$) on these same planets, confirming that the correct physical world model exists in the backbone but is destroyed by the invasive measurement.

\section{Parameter Change Analysis}
\label{app:paramchange}

To quantify the invasiveness of the fine-tuning process, we analyze the magnitude of weight modifications across the transformer architecture. We compute the layer-wise relative change using the Frobenius norm:
\begin{equation}
    \delta^{(l)} = \frac{\|\theta'^{(l)} - \theta^{(l)}\|_F}{\|\theta^{(l)}\|_F}
\end{equation}
where $\theta^{(l)}$ represents the weights of layer $l$ in the pre-trained SSL model, and $\theta'^{(l)}$ represents the weights after fine-tuning on the inductive bias task.

\begin{figure}[ht]
    \centering
        \includegraphics[width=\linewidth]{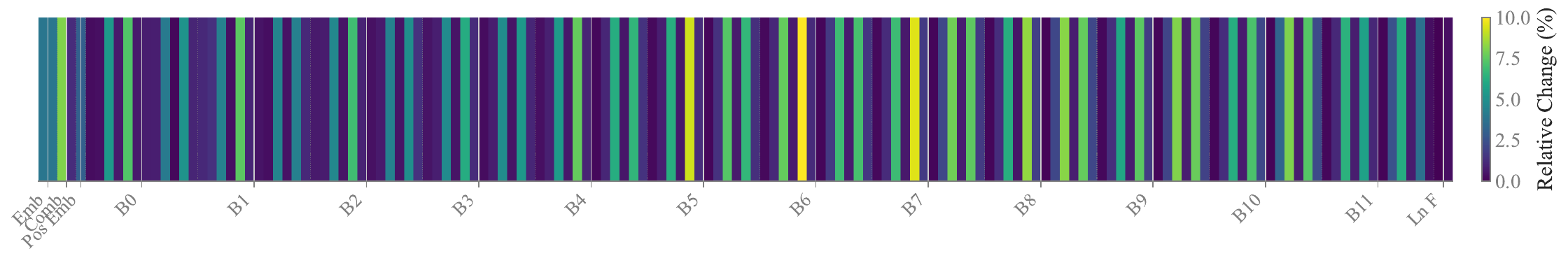}
        \caption{\textbf{Global Invasiveness:} Parameter heatmap showing modification concentrated in deep layers.}
        \label{fig:param_change}
\end{figure}
As illustrated in Figure \ref{fig:param_change}, the modification is highly non-uniform.
\begin{itemize}
    \item \textbf{Early Layers (Blocks 0--4):} Exhibit high stability ($\delta^{(l)} < 0.02$), indicating that the basic feature extraction for orbital trajectories remains largely intact.
    \item \textbf{Deep Layers (Blocks 5--10):} Show a sharp spike in parameter modification ($\delta^{(l)} > 0.10$), particularly in the MLP projections and Self-Attention output matrices.
\end{itemize}
This concentrated modification in the deeper blocks corroborates our ``Erasure'' hypothesis. In hierarchical models, deep layers typically encode high-level semantic variables and dynamic rules \citep{tenney2019bert}. The fact that the optimizer selectively targets these layers suggests it is overwriting the model's high-level physics engine (the ``World Model'') to replace it with the shallow heuristics required by the narrow fine-tuning distribution.
% \section{Parameter change analysis after finetuning}
% \label{app:paramchange}

% \begin{figure}[ht]
%     \centering
%         \includegraphics[width=\linewidth]{Figures/changeanalysis (1).pdf}
%         \caption{\textbf{Global Invasiveness:} Parameter heatmap showing modification concentrated in deep layers.}
%         \label{fig:param_change}
% \end{figure}
\section{Full Fine-tuning Data Distribution Analysis}
\label{app:ftdataan}

We analyze the distributional shift between the self-supervised pre-training dataset ($D_{\text{SSL}}$) and the downstream fine-tuning dataset ($D_{\text{task}}$).

\begin{figure}[ht]
    \centering
    % Ensure this filename points to your HISTOGRAM plot (Star Mass/Force Dist), NOT the heatmap
    \includegraphics[width=\linewidth]{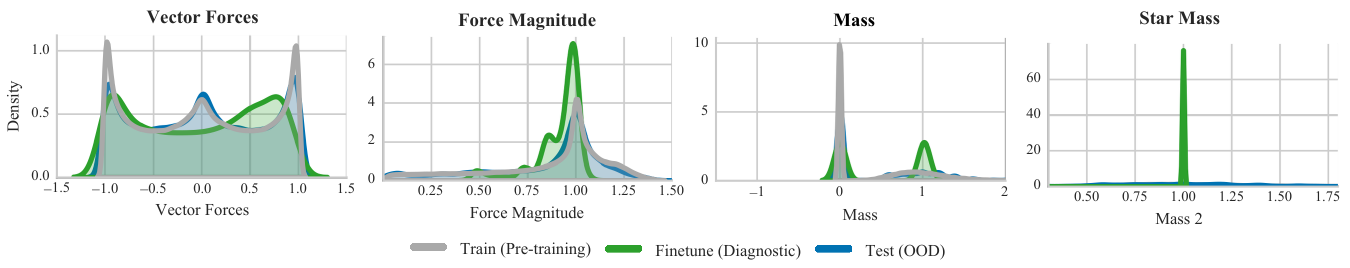} 
    \caption{\textbf{Data Distribution Shift.} Comparison of physical variables between SSL (Grey) and Fine-tuning (Green). The fine-tuning data collapses to narrow regimes (e.g., single star mass), inducing simplicity bias.}
    \label{fig:data_dist_full}
\end{figure}

As hypothesized in \Cref{sec:finetuning}, invasive adaptation on narrow distributions encourages the model to discard complex physical dependencies in favor of statistical shortcuts. \Cref{fig:data_dist_full} illustrates this discrepancy across key variables:

\begin{itemize}
    \item \textbf{Star Mass ($m_2$):} While the SSL pre-training data covers a continuous range of stellar masses, the fine-tuning dataset collapses to a single point mass at $m_2 = 1.0$. This lack of variance removes the incentive for the model to maintain $m_2$ as an active variable, leading to the ``erasure'' observed in our mechanistic analysis.
    
    \item \textbf{Force Magnitude ($||\vec{F}||$):} The fine-tuning dataset exhibits a highly peaked distribution centered around $||\vec{F}|| \approx 1.0$, failing to cover the heavy tails of high-force interactions  or low-force interactions. This restricts the optimizer's ability to learn the full inverse-square law~($1/r^2$), as gradients are dominated by a specific force regime.
    
    \item \textbf{Force Vectors ($\vec{F}$):} The distribution of vector components in the fine-tuning set differs significantly from the isotropic distribution seen during pre-training, potentially overfitting the model to specific orbital orientations rather than learning rotation-invariant physics.
\end{itemize}

\section{Symbolic Formula Comparison}
\label{app:symbolic_comparison}

Table \ref{tab:symbolic_comparison_all} presents the symbolic equations discovered for the gravitational force magnitude ($F$) via Symbolic Regression (SR). We compare the equations extracted from the Invasive Fine-Tuning baseline (IBP)  \cite{vafa2025foundationmodelfoundusing} against those recovered by \textit{PhyIP}.
The SR algorithm (PySR) attempts to fit the scalar force magnitude $\|\vec{F}\|$ using the state variables $r$ (distance), $m_1$ (planet mass), and $m_2$ (star mass). The search was constrained to standard arithmetic and trigonometric (+, -, *, /, sin, cos)

\begin{table}[h]
\centering
\small  % Sets a clean, readable font size without stretching
\caption{\textbf{Symbolic Equation Discovery.} Comparison of discovered laws. The Invasive Probe fits spurious correlations (nested sines), while the Non-Invasive Probe recovers the structural $1/r^2$ dependence.}
\label{tab:symbolic_comparison_all}
\renewcommand{\arraystretch}{1.5} % Good spacing, but not too tall
\begin{tabular}{l l l}
\toprule
\textbf{Source} & \textbf{Complexity} & \textbf{Discovered Symbolic Equation} \\
\midrule

\rowcolor{gray!10}
\textbf{Ground Truth} & -- & $\displaystyle \|\vec{F}\| \propto \frac{m_1 m_2}{r^2}$ \\

\midrule

\textbf{Invasive FT (IBP)} & High & $\displaystyle \|\vec{F}\| \propto \left[\sin\left(\frac{1}{\sin(r-0.24)}\right) + 1.45\right] \cdot \frac{1}{1/r + m_2}$ \\
\textit{(Baseline)} & & \textit{(Fails to isolate $1/r^2$; relies on high-freq artifacts)} \\

\midrule

\textbf{Non-Invasive (Ours)} & Low (Rank 1) & $\displaystyle \|\vec{F}\| \approx \frac{1}{1/r + 1.16}$ \\

\textbf{Non-Invasive (Ours)} & Med (Rank 2) & $\displaystyle \|\vec{F}\| \approx \sin(\sin(\sin(r \cdot 0.07) + 0.48))$ \\

\textbf{Non-Invasive (Ours)} & \textbf{Best (Rank 3)} & $\displaystyle \|\vec{F}\| \approx \underbrace{\frac{\mathbf{0.1}}{\mathbf{r^2}}}_{\text{Recovered Physics}} + \underbrace{\sin(\dots)}_{\text{Residual Noise}}$ \\

\bottomrule
\end{tabular}
\end{table}

% \begin{table}[ht]
% \centering
% \caption{Comparison of Symbolic Equations for Force Magnitude ($\|\vec{F}\|$) derived via SR.}
% \label{tab:symbolic_comparison_all}
% \resizebox{\linewidth}{!}{% Resize table to fit within line width
% \begin{tabular}{@{}lll@{}}
% \toprule
% Method & Equation Name & Symbolic Equation \\
% \midrule
% \textbf{Ground Truth} & \textbf{Newton's Law} & $\mathbf{\|\vec{F}\| \propto \frac{m_1 m_2}{r^2}}$ \\
% \addlinespace % Adds a bit of vertical space

% SR on Fine-Tuning & \citep{vafa2025foundationmodelfoundusing} Eq. & $\|\vec{F}\| \propto \left(\sin\left(\frac{1}{\sin(r-0.24)}\right) + 1.45\right) \cdot \frac{1}{1/r + m_2}$ \\
% \addlinespace % Adds a bit of vertical space

% SR on Probe & Eq 1 & $\|\vec{F}\| \approx \frac{1}{1/r + 1.1601468}$ \\
% \addlinespace

% SR on Probe & Eq 4 & $\|\vec{F}\| \approx \sin(\sin(\sin(r \cdot 0.072632626) + 0.48367104))$ \\
% \addlinespace

% SR on Probe & Eq 5 & $\|\vec{F}\| \approx \cos(\cos(\sin(\frac{1}{r \cdot 0.14463314} + 0.20845711)))$ \\
% \addlinespace

% SR on Probe & Eq 6 & $\|\vec{F}\| \approx \sin(\sin((r \cdot 0.03789409) + \sin(m_2 \cdot 0.06399643) + 0.44295916))$ \\
% \addlinespace

% SR on Probe & Eq 7 & $\|\vec{F}\| \approx \sin(\sin(\sin(((m_2^2 + r) \cdot (m_2 \cdot 0.062196407)) + 0.38174748) + (r \cdot 0.038285665)))$ \\
% \addlinespace

% SR on Probe & Eq 7 & $\begin{aligned}[t] \|\vec{F}\| \approx & \sin(\sin(\sin(((m_2^2 + r) \cdot (m_2 \cdot 0.062196407)) + 0.38174748) + (r \cdot 0.038285665))) \\ & + \frac{0.1}{r^2} \end{aligned}$ \\
% \bottomrule
% \end{tabular}%
% } % End resizebox
% \end{table}

To ensure reproducibility, we generated 7 candidate equations using PySR's simulated annealing. The ``Best'' equation reported in Table \ref{tab:symbolic_comparison_all} was selected using the Pareto frontier of the \texttt{score} metric (minimizing MSE while penalizing complexity).

\end{document}